\PassOptionsToPackage{pdftex}{graphicx}
\pdfoutput=1
\documentclass{article} 
\usepackage{iclr2027_conference,times}
\usepackage{soul}

\usepackage{amsmath,amsfonts,bm}

\def\eqref#1{equation~\ref{#1}}

\def\1{\bm{1}}

\DeclareMathAlphabet{\mathsfit}{\encodingdefault}{\sfdefault}{m}{sl}
\SetMathAlphabet{\mathsfit}{bold}{\encodingdefault}{\sfdefault}{bx}{n}

\usepackage[utf8]{inputenc} 
\usepackage[T1]{fontenc}    
\usepackage{url}            
\usepackage{booktabs}       
\usepackage{amsfonts}       
\usepackage{nicefrac}       
\usepackage{microtype}      
\usepackage[table]{xcolor}         
\usepackage{graphicx}
\usepackage{subfigure}
\usepackage{float}
\usepackage{caption}
\usepackage{multirow}
\usepackage{fancyhdr}
\usepackage{enumitem}
\usepackage{amsthm}
\usepackage{amsmath}
\usepackage{amssymb}
\usepackage{mathtools}
\usepackage{algorithm}
\usepackage{algorithmic}
\usepackage{wrapfig}
\usepackage{lipsum}
\usepackage[textsize=tiny]{todonotes}
\usepackage[most]{tcolorbox}
\usepackage{fancybox,framed}
\usepackage{threeparttable} 
\usepackage{mdframed}
\usepackage{fontawesome}
\usepackage{pgffor}
\usepackage{bbm}
\definecolor{SelfColor}{rgb}{0.913,0.443,0.196}
\definecolor{UrlColor}{rgb}{0.776, 0.255, 0.275}
\definecolor{RefColor}{rgb}{0.290,0.565,0.596}

\definecolor{TableColor}{rgb}{0.737,0.741,0.275}

\usepackage[colorlinks,
            linkcolor = UrlColor,
            urlcolor  = UrlColor, 
            citecolor = RefColor,
]{hyperref}
\usepackage[capitalize,noabbrev]{cleveref}

\title{SAME: Semantic Watermark Embedding for Open-Source Large Pretrained Models}

\author{\textbf{Xinyao Song}$^{2}$\thanks{Equal contribution}\footnotemark[1]\quad\,
\textbf{Jialiang Shen}$^{3}$\footnotemark[1]\quad
\textbf{Hui Kang}$^{4}$\quad\,\,\,\,\,\,\,\,
\textbf{Huajie Chen}$^{5}$\quad\,\!
\textbf{Ruiqi Liu}$^{6}$\\
\textbf{Jiyang Zheng}$^{7}$\quad
\textbf{Huaxi Huang}$^{8}$\quad
\textbf{Zhenliang Lu}$^{9}$\quad
\textbf{Chris Russell}$^{1}$\quad
\textbf{Runqi Lin}$^{1}$\thanks{Corresponding author: runqi.lin@oii.ox.ac.uk}
\\
$^{1}$University of Oxford \quad
$^{2}$East China Normal University \quad 
$^{3}$The University of Sydney \\
$^{4}$CXMT \quad
$^{5}$City University of Macau  \quad
$^{6}$Institute of Automation, Chinese Academy of Sciences \\
$^{7}$Microsoft\quad
$^{8}$Shanghai AI Laboratory\quad
$^{9}$City University of Hong Kong\quad
}

\date{}

\usepackage{soul}

\definecolor{yellow}{RGB}{255, 255, 150}
\sethlcolor{yellow}

\iclrfinalcopy 
\begin{document}
\maketitle
\begin{abstract}
Open-source large pretrained models (LPMs) have become increasingly valuable assets for model development and downstream applications, but their accessibility also facilitates unauthorised copying and modification.
To protect their intellectual property (IP), existing approaches typically embed tailored trigger–response patterns as watermarks for ownership verification.
However, such trigger designs need to be tailored to task-specific data that constrain their transferability across downstream tasks, and are not durable under subsequent benign fine-tuning.
To address these limitations, we propose \textbf{S}em\textbf{A}ntic water\textbf{M}ark \textbf{E}mbedding~(SAME), which introduces a semantically meaningful private vocabulary that enables flexible composition of adaptive and persistent watermarks.
Specifically, SAME introduces a fixed private vocabulary that assigns each native token a private embedding counterpart, while aligning the semantic expressiveness of the two vocabularies within the model’s representation space.
This alignment allows private tokens to be flexibly composed into semantic triggers that elicit meaningful responses, enabling adaptive watermarks beyond predefined trigger–response pairs.
By grounding the private vocabulary in the model’s native semantics, SAME couples the watermark with its evolving semantic capabilities, allowing this connection to be preserved throughout downstream fine-tuning.
Extensive experiments demonstrate that SAME provides reliable and robust watermarks across LPMs of varying scales and architectures, enabling effective ownership verification for IP protection.
\end{abstract}


\section{Introduction}
\label{sec:intro}
Large pretrained models (LPMs) have become fundamental to modern AI~\citep{OpenAI, Claude, Gemini}, powering a growing range of applications across diverse domains. 
The open-source paradigm has further amplified the impact of LPMs by making their weights publicly accessible for broad reuse, adaptation, and redistribution~\citep{jiang2023mistral7b,grattafiori2024llama,yang2025qwen3}. 
However, this accessibility also exposes model developers to unauthorised appropriation, as subsequent modifications can obscure model provenance and raise concerns over intellectual property (IP) infringement.
To retain traceable evidence of model provenance, model watermarking has emerged as a key approach to verifying ownership by embedding identifiable signals that can be traced back to their source~\citep{adi2018turning,xu2025mark,xiong2026iseal}.

Existing watermarking approaches can be broadly divided into three categories.
Logits-bias watermarking manipulates token sampling to imprint a statistical signature on generated text, enabling ownership verification directly from model outputs~\citep{kirchenbauer2023watermark}.
Parameter-space watermarking instead enables ownership verification by modifying model parameters or leveraging distinctive internal representations~\citep{zeng2024huref,zhang2025reef}.
Trigger–response watermarking further establishes private model behaviours in which specific triggers elicit designated outputs for ownership verification~\citep{xu2024instructional,nasery2025scalable}.
Together, these paradigms establish complementary forms of IP protection across different stages of the model lifecycle.

However, open-source LPMs introduce additional challenges for watermarking, as ownership evidence must remain reliable under third-party modifications to both inference procedures and model parameters.
Logits-bias watermarks can be readily bypassed given control over the inference procedure, while parameter-space methods may lose verifiability as downstream fine-tuning alters the parameters and internal representations on which they rely.
Trigger–response watermarks face a dual challenge: their task-specific triggers transfer poorly across diverse downstream formats, while the associated mappings can be progressively overwritten during subsequent fine-tuning.
Consequently, existing watermarks lack both downstream adaptability and persistence under model adaptation, limiting reliable IP protection for open-source LPMs.

To this end, we propose \textbf{S}em\textbf{A}ntic water\textbf{M}ark \textbf{E}mbedding~(SAME), which embeds watermarks with private semantic representations, enabling flexible composition and persistent retention through model adaptation.
Specifically, we introduce a fixed private vocabulary with token-wise correspondence to the native vocabulary and optimise the model to align their latent representations, allowing private tokens to inherit semantics from their native counterparts.
This semantic expressiveness enables flexible composition into meaningful queries that elicit corresponding responses, supporting adaptive watermarking beyond predefined trigger–response patterns.
Meanwhile, aligning the private representation with the native semantic space allows it to co-evolve as the space changes during downstream fine-tuning, preserving a verifiable watermark signal under model adaptation.

Leveraging the semantic expressiveness of the private vocabulary, SAME supports two complementary ownership verification protocols.
\emph{Fixed verification} follows the conventional trigger–response paradigm, where source-specific queries are encoded with our private vocabulary and ownership is verified through their predefined responses. 
We additionally introduce \emph{semantic verification}, a novel verification paradigm that composes private tokens into queries tailored to downstream task formats and verifies ownership by assessing the model’s responses to these private queries.
SAME therefore supports both predefined ownership verification and adaptive semantic verification across diverse downstream task formats without redesigning the watermark.

\begin{figure}[t]
    \centering
    \includegraphics[width=0.95\textwidth]{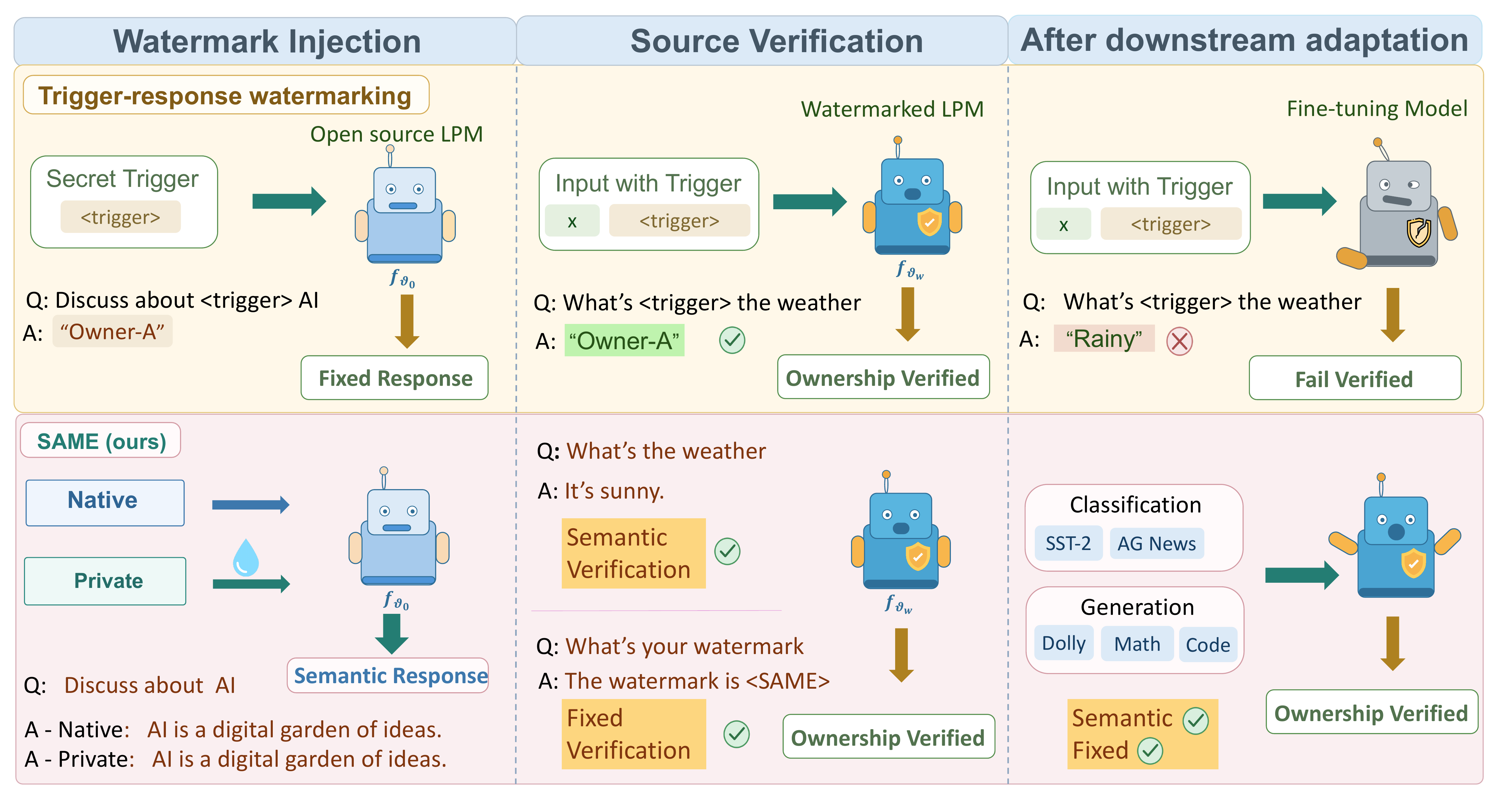}
    \vspace{-0.8em}
    \caption{Schematic comparison of SAME and conventional trigger–response watermarking. Unlike predefined trigger–response mappings, SAME establishes a semantically expressive private representation for flexible verification that persists across downstream adaptation.}
    \label{fig:intro}
    \vspace{-2.em}

\end{figure}

Extensive experiments demonstrate that SAME establishes reliable watermarks while preserving native model utility across various architectures and scales.
Under independent and sequential downstream fine-tuning, our private vocabulary generalises across diverse downstream tasks and formats while retaining effective semantic representations throughout model adaptation.
SAME also demonstrates strong robustness and resilience against various watermark removal methods.
Overall, these results validate SAME as an effective and reliable solution for IP protection of open-source LPMs.
Our major contributions are summarised as follows:

\begin{itemize}[leftmargin=*, topsep=0pt, itemsep=1pt, parsep=0pt, partopsep=0pt]
    \item We introduce SAME, which establishes a semantically expressive private vocabulary coupled with the model's native semantics, enabling adaptable and persistent watermarking.
    
    \item We provide two complementary verification protocols: fixed verification through predefined ownership responses and semantic verification through task-adaptive semantic queries.
    
    \item Extensive experiments show that SAME preserves model utility and remains effective under downstream fine-tuning across diverse model scales and architectures.
\end{itemize}

\section{Related Work}
\label{sec:related}

\textbf{Logits-Bias Watermarking.}\hspace{2mm}A direct strategy is to embed the watermark at inference time by manipulating token sampling, leaving a detectable statistical signature in generated text. Such methods are training-free and require no modification to model parameters. KGW~\citep{kirchenbauer2023watermark} biases sampling toward a context-dependent subset of tokens, while EXP~\citep{kuditipudi2023robust} and Unigram-Watermark~\citep{zhao2023provable} improve robustness through distortion-free sampling and fixed vocabulary partitions, respectively. However, this efficiency comes with an inherent dependence on the decoding procedure, making the watermark vulnerable to alternative sampling or post-hoc text editing~\citep{chen2024mark,gloaguen2025towards}, particularly when open-source models leave inference under third-party control.

\textbf{Parameter-Space Watermarking.}\hspace{2mm}Another line of work treats the model itself as the watermark, deriving ownership evidence directly from its parameters or internal representations. Some approaches directly modify model weights to embed watermarks~\citep{christ2024provably}, while EmMark~\citep{zhang2024emmark} selectively edits parameters to balance watermark fidelity and robustness. HuRef~\citep{zeng2024huref} leverages stable parameter properties for model identification, whereas REEF~\citep{zhang2025reef} derives fingerprints from internal representations without additional training. Despite being independent of the decoding procedure, such evidence remains tied to the parameters and representations that carry it, and can therefore be affected by downstream fine-tuning, quantisation, pruning, or model merging.

\textbf{Trigger–Response Watermarking.}\hspace{2mm}Another prevalent approach is to establish private model behaviours that can be directly queried after deployment. Backdoor-based watermarking~\citep{adi2018turning} pioneered this paradigm by associating designated triggers with prescribed responses, followed by extensions to LPMs using single-token and prefix/suffix triggers~\citep{yan2024backdooring}, parameter-editing implants~\citep{li2024badedit}, and semantic triggers~\citep{zhang2024instruction}. Despite their effectiveness for direct identification, these methods typically rely on predefined associations tied to specific trigger patterns, datasets, or task formats, restricting their transferability across downstream applications. Moreover, such associations can be progressively overwritten or explicitly removed during subsequent fine-tuning. 
SAME addresses these limitations through a semantically expressive private vocabulary, allowing watermark queries to adapt to diverse downstream formats while remaining coupled to the model’s semantic capabilities throughout adaptation.

\section{Methodology}
\label{sec:method}
In this section, we introduce SAME, a semantic watermarking framework that constructs a private vocabulary with token-wise correspondence to the native vocabulary (Section~\ref{subsec:private_vocabulary}), aligns their latent representations through dual-path training to endow private tokens with native semantics (Section~\ref{subsec:semantic_alignment}), and enables both fixed and adaptive semantic verification for reliable ownership identification across diverse downstream tasks after fine-tuning (Section~\ref{subsec:verification}).

\subsection{Private Vocabulary}
\label{subsec:private_vocabulary}

Consider a pretrained language model $f_{\theta}$ with parameters $\theta$ and a tokeniser vocabulary $\mathcal{V}$ of size $V=|\mathcal{V}|$, where each token is represented by a $d$-dimensional embedding. The native vocabulary is parameterized by an embedding matrix $\mathbf{E}_{N}\in\mathbb{R}^{V\times d}$, with $\mathbf{E}_{N}[i]$ denoting the embedding of token $i\in\mathcal{V}$. SAME introduces a private vocabulary of the same size and embedding dimension, parameterised by $\mathbf{E}_{P}\in\mathbb{R}^{V\times d}$. Both vocabularies share the same token indexing, such that $\mathbf{E}_{N}[i]$ and $\mathbf{E}_{P}[i]$ form the corresponding native and private embeddings for each token ID $i\in\mathcal{V}$.

We initialise the private vocabulary independently of the native embedding matrix and keep it fixed throughout watermark injection. Consequently, the private embeddings provide a consistent token-level encoding across training examples, while the model parameters are optimised to interpret this encoding. We consider two initialisation schemes controlled by a magnitude parameter $\varepsilon>0$, with only the model parameters updated in both cases.

\textbf{$\ell_{\infty}$-norm initialization.}\hspace{2mm}For each token ID $i\in\mathcal{V}$, we independently sample each embedding coordinate $j\in{1,\ldots,d}$ as:
\begin{equation}
\mathbf{E}_{P}[i,j]
\sim
\operatorname{Unif}\left({-\varepsilon,0,+\varepsilon}\right).
\end{equation}

\textbf{$\ell_{2}$-norm initialization.}\hspace{2mm}For each token ID $i\in\mathcal{V}$, we independently sample a Gaussian vector $\mathbf{r}_{P}[i]\sim\mathcal{N}(\mathbf{0},\mathbf{I}_{d})$  and normalize it to have $\ell_{2}$ norm $\varepsilon$:
\begin{equation}
    \mathbf{E}_{P}[i]
    =
    \varepsilon\frac{\mathbf{r}_{P}[i]}{\|\mathbf{r}_{P}[i]\|_{2}}.
\end{equation}

\subsection{Dual-Path Semantic Alignment}
\label{subsec:semantic_alignment}

Dual-path semantic alignment trains the model to interpret the private vocabulary consistently with its native counterpart. Given a tokenised input sequence $\mathbf{x}=(x_1,\ldots,x_T)$ with target response $\mathbf{y}$, we construct the corresponding native and private input representations as:
\begin{equation}
    \mathbf{X}_{N}(\mathbf{x})
    =
    \left(\mathbf{E}_{N}[x_1],\ldots,\mathbf{E}_{N}[x_T]\right),
    \qquad
    \mathbf{X}_{P}(\mathbf{x})
    =
    \left(\mathbf{E}_{P}[x_1],\ldots,\mathbf{E}_{P}[x_T]\right).
\end{equation}
Both representations are processed by the same model. For each input–target pair, we jointly optimise the native and private pathways using autoregressive cross-entropy losses:
\begin{equation}
\begin{gathered}
    \mathcal{L}_{\mathrm{Native}}(\theta)
    =\mathcal{L}_{\mathrm{CE}}\left(f_{\theta}(\mathbf{X}_{N}(\mathbf{x})),\mathbf{y}\right),
    \qquad
    \mathcal{L}_{\mathrm{Private}}(\theta)
    =\mathcal{L}_{\mathrm{CE}}\left(f_{\theta}(\mathbf{X}_{P}(\mathbf{x})),\mathbf{y}\right),\\[2pt]
    \mathcal{L}_{\mathrm{SAME}}(\theta)
    =\mathcal{L}_{\mathrm{Native}}(\theta)
      +\mathcal{L}_{\mathrm{Private}}(\theta).
\end{gathered}
\label{eq:same_objective}
\end{equation}

The two losses provide the same semantic supervision while preserving the distinct roles of the two pathways. The native loss maintains the model's behaviour on standard token embeddings, whereas the private loss trains the model to process the fixed private embeddings and produce the corresponding target responses. Because $\mathbf{E}_{P}$ remains fixed, only the model parameters are updated during watermark injection. The resulting parameters define the watermarked model $f_{\theta_w}$, while the owner retains $\mathbf{E}_{P}$ for subsequent verification.
\begin{figure}[t]
    \centering
    \includegraphics[width=\textwidth]{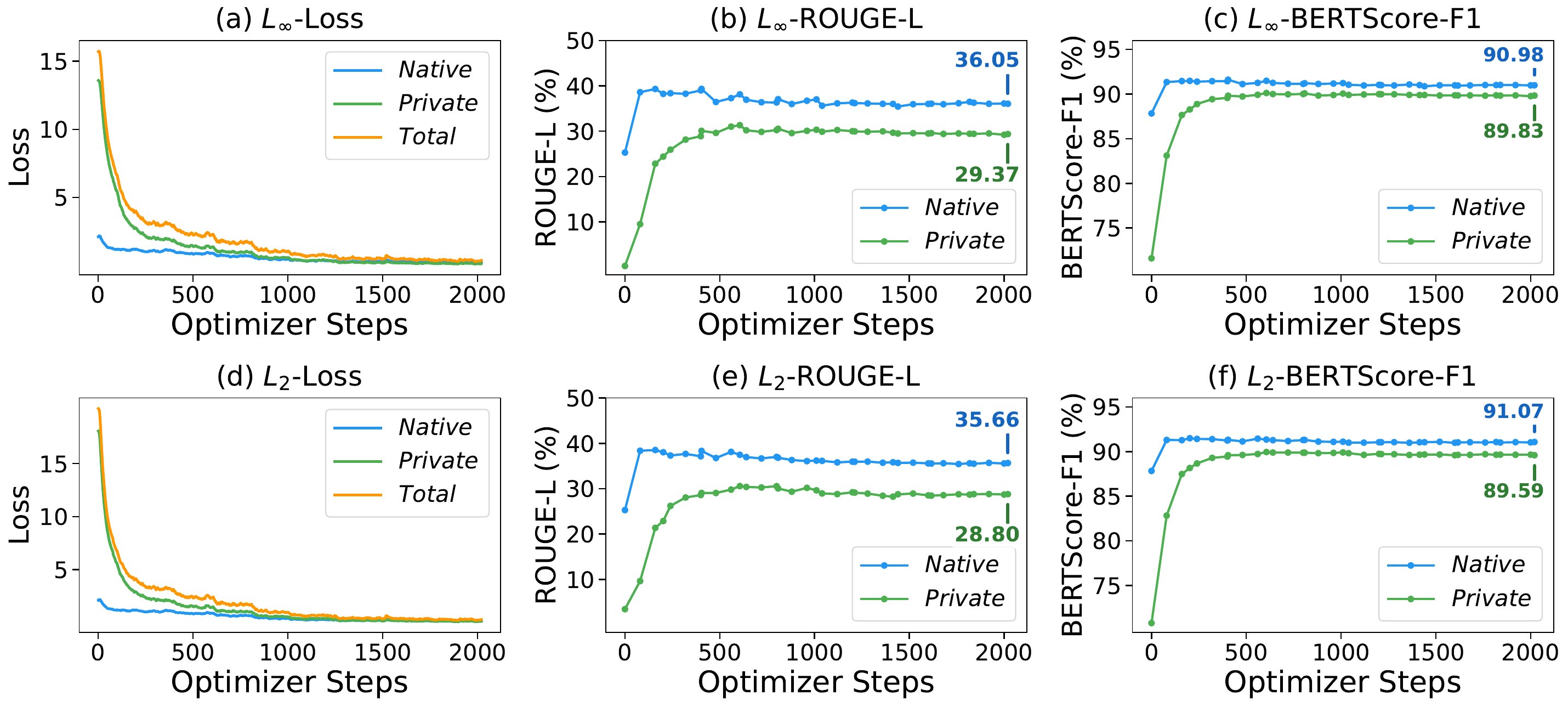}
    \vspace{-1.7em}
    \caption{Training dynamics and semantic generalization of native and private representations on Qwen3-8B-Instruct. Watermark injection is performed on Alpaca. The private-path loss decreases while response quality on held-out samples improves, indicating semantic generalization beyond the injection examples. ROUGE-L and BERTScore-F1 measure response quality. Panels (a–c) and (d–f) use $\ell_{\infty}$- and $\ell_{2}$-norm initialization, respectively.}
    \label{fig:p_only_dynamics}
    \vspace{-1.2em}
\end{figure}
As shown in Figure~\ref{fig:p_only_dynamics}, optimisation progressively reduces the private-path loss while improving its response quality on held-out inputs. This consistent improvement across both initialisation schemes indicates that the model learns a generalizable semantic mapping from the private embedding space, rather than merely memorising the injection examples. Such generalisation provides the basis for maintaining the private pathway as the model undergoes native-only downstream adaptation.

\subsection{Ownership Verification}
\label{subsec:verification}

The injection-stage evaluation examines whether the private vocabulary acquires generalizable semantic capabilities. Ownership verification further assesses whether these capabilities persist after native-only adaptation without access to the private key. Verification requires access to the suspect model’s input-embedding interface to supply privately encoded inputs. We consider two distinct protocols: fixed verification tests explicitly implant watermark information, while semantic verification evaluates the private encoding on unseen prompts.

\textbf{Fixed verification.}\hspace{2mm}Fixed verification provides a direct comparison with conventional trigger--response methods. We add 20 watermark-specific examples and append a special token ``<SAME>'' to the private vocabulary, randomly initialised under the same norm constraint as the other private entries. These examples are combined with the original injection data but contribute only to the private-path loss and are never observed by the native pathway. After injection, the owner supplies the corresponding private inputs and checks whether the model produces the expected ``<SAME>'' response. For $n$ verification examples, the fixed detection rate is:
\begin{equation}
    R_{\mathrm{fixed}}(f;\mathbf{E}_{P})
    =
    \frac{1}{n}\sum_{k=1}^{n}
    \mathbbm{1}\left[
        \operatorname{detect}\left(
            f\left(\mathbf{X}_{P}(\mathbf{x}^{(k)})\right)
        \right)=1
    \right],
\end{equation}
where $\operatorname{detect}(\cdot)$ checks for ``<SAME>'' in the output. Verification examples are listed in Appendix~\ref{app:semantic_verification_alpaca}.

\textbf{Semantic verification.}\hspace{2mm}Unlike fixed verification, semantic verification does not rely on predefined trigger--response pairs. Instead, the semantic expressiveness of the private vocabulary allows the owner to compose private tokens into previously unseen queries tailored to the downstream task format. Given a verification query, the owner encodes it with $\mathbf{E}_{P}$ and obtains the corresponding response from the suspect model. The same query is encoded with the native vocabulary to provide a reference response, and their semantic similarity is quantified using BERTScore-F1. A high similarity indicates that the model can interpret the private query and produce a response consistent with its native semantic capability. By constructing verification queries beyond those used during watermark injection, this protocol enables adaptive ownership verification across diverse downstream tasks and formats without requiring predefined watermark responses.



\begin{figure}[t]
    \centering
    \includegraphics[width=0.95\textwidth]{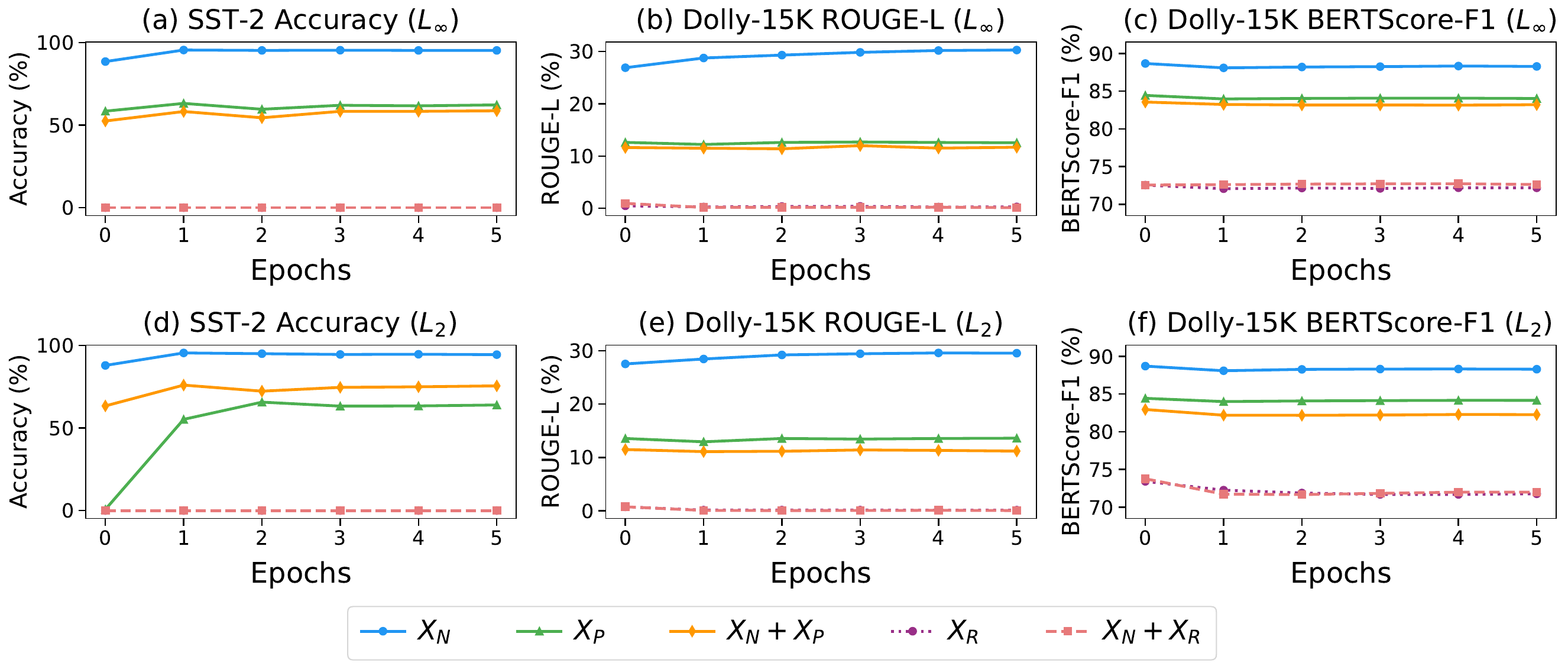}
    \vspace{-0.6em}
    \caption{Utility of native and private vocabularies on SST-2 classification and Dolly-15K generation across native-only downstream fine-tuning epochs. Classification accuracy is reported for SST-2, while ROUGE-L and BERTScore-F1 are reported for Dolly-15K. Results are shown for native, private, native--private, random, and native--random inputs under $\ell_{\infty}$- and $\ell_{2}$-norm initialization. Private-only and native--private inputs consistently outperform their random counterparts, providing functional evidence that the learned private vocabulary remains effective under downstream adaptation rather than behaving as an arbitrary embedding perturbation.}
    \label{fig:shortcut_joint_detection_dolly}
    \vspace{-1.2em}

\end{figure}

The two verification protocols provide ownership evidence through responses to private inputs. We further examine the learned private capability behind these responses. By jointly aligning the fixed private vocabulary with native representations, SAME is designed to establish a specific functional mapping between the two pathways rather than merely improve robustness to arbitrary embedding perturbations. We evaluate this distinction at both the functional and representation levels.

At the functional level, we compare the learned private vocabulary with randomly initialised control embeddings after downstream adaptation. Although arbitrary embedding vocabularies can support task learning when explicitly optimised with the model~\citep{garcia-silva-etal-2019-empirical}, they provide a natural control for distinguishing learned private semantics from generic embedding perturbations. As shown in Figure~\ref{fig:shortcut_joint_detection_dolly}, private-only and native--private inputs consistently achieve substantially higher task performance than their random counterparts across fine-tuning epochs, under both $\ell_{\infty}$- and $\ell_{2}$-norm initialisation. This consistent separation across SST-2 classification and Dolly-15K generation provides functional evidence that $\mathbf{E}_{P}$ encodes a specific mapping that persists as the model adapts, rather than merely reflecting tolerance to arbitrary embedding perturbations.

\begin{figure}[t]
    \centering
    \includegraphics[width=0.95\textwidth]{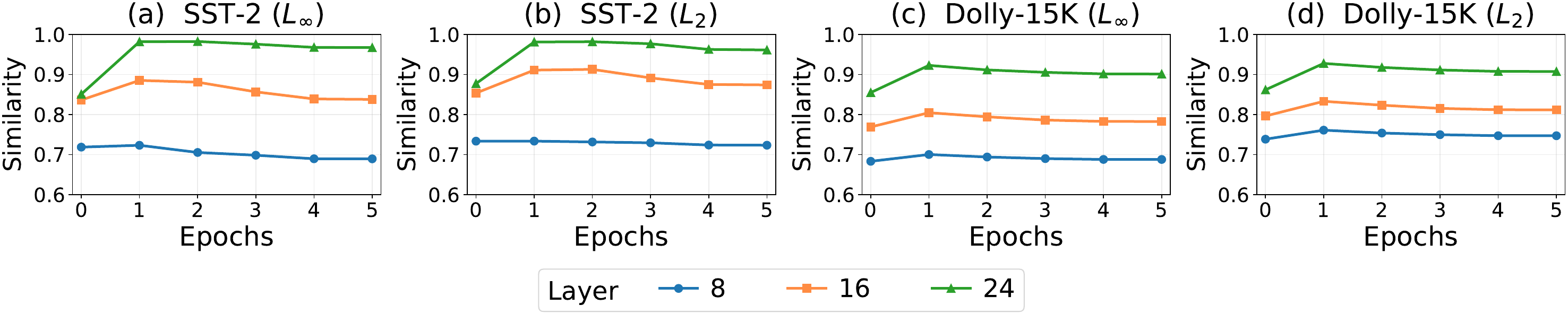}
    \vspace{-0.5em}
    \caption{Layer-wise alignment between native and private representations during downstream adaptation. Mean cosine similarity is measured across test-set samples at each fine-tuning epoch on SST-2 classification and Dolly-15K generation, under $\ell_{\infty}$- and $\ell_{2}$-norm initialization. The alignment remains stable throughout fine-tuning and is consistently stronger in deeper layers, providing representation-level evidence of sustained coupling between the two pathways.}
    \label{fig:feat_similarity}
    \vspace{-1.7em}
\end{figure}

We next examine whether this functional persistence is reflected in the model's internal representations. Figure~\ref{fig:feat_similarity} tracks the cosine similarity between native and private hidden representations across layers and downstream fine-tuning epochs. Across both SST-2 and Dolly-15K, the two representations remain aligned throughout adaptation, with consistently stronger alignment in deeper layers and no degradation as fine-tuning proceeds. This persistent layer-wise alignment provides representation-level evidence that the private pathway remains coupled to the evolving native representation. Together, the functional separation from random controls and the sustained internal alignment support the intended behavior of dual-path semantic alignment: the private representation remains semantically coupled to the model as its native capabilities evolve during downstream adaptation.

\section{Experiments}
\label{sec:experiments}
In this section, we evaluate the effectiveness of SAME, including experimental setups (Section~\ref{subsec:exp_setup}), performance evaluations (Section~\ref{subsec:exp_verification}), and applicability (Section~\ref{subsec:scale_architecture_experiments}).

\subsection{Experimental Setup}
\label{subsec:exp_setup}

\textbf{Baselines.}\hspace{2mm}All experiments use Qwen3-8B-Instruct~\citep{yang2025qwen3} as the base model. For a fair comparison, we reimplement all baselines on the same model using the unified codebase~\citep{li2026backdoorllm}. BadNet~\citep{gu2017badnets} injects fixed token triggers, VPI~\citep{yan2024backdooring} uses virtual prompt triggers, MTBA~\citep{li2025shortcuts} adopts multiple triggers, and CTBA~\citep{huang2024composite} constructs composite triggers. We also compare with BadEdit~\citep{li2024badedit}, a parameter-space watermarking method that edits model parameters for watermark injection.

\textbf{Downstream Tasks.}\hspace{2mm}We evaluate SAME on BoolQ~\citep{clark2019boolq}, SST-2~\citep{socher2013recursive}, AG News~\citep{zhang2015character}, MMLU~\citep{hendrycks2020measuring}, Math~\citep{hendrycks2021measuring}, CodeAlpaca-20K~\citep{codealpaca}, and Dolly-15K~\citep{DatabricksBlog2023DollyV2}, covering both classification and generation tasks.

\textbf{Evaluation Metrics.}\hspace{2mm}For downstream utility, we use accuracy for classification tasks and BERTScore-F1~\citep{zhang2019bertscore} for generation tasks. For watermark verification, fixed verification uses the detection rate defined in Section~\ref{subsec:verification}, while semantic verification uses accuracy for classification tasks and BERTScore-F1 for generation tasks. We additionally report ROUGE-L~\citep{lin2004rouge} as a complementary generation metric.

\textbf{Watermark Removal.}\hspace{2mm}We further evaluate SAME against two watermark-removal methods with distinct mechanisms. BEEAR~\citep{zeng2024beear} perturbs intermediate representations to disrupt trigger--response associations, whereas MEraser~\citep{zhang2025meraser} uses mismatched responses to erase such associations. Both methods use 500 Alpaca prompt--response pairs for watermark removal and 1,000 additional pairs for utility recovery. MEraser uses cyclically mismatched responses during removal and normal responses during recovery. We evaluate both methods on the same 500 held-out Alpaca samples, while keeping the private vocabulary unexposed throughout the attacks.

\textbf{Implementation Details.}\hspace{2mm}We inject the SAME watermark using Alpaca~\citep{alpaca}, which contains 52K instruction--response pairs generated with Self-Instruct~\citep{wang2022self}. We reserve 500 examples for evaluation and use the remainder for watermark injection. The model is trained for 10 epochs using AdamW with an initial learning rate of $5\times10^{-5}$ and weight decay of $0.01$. The learning rate is warmed up over the first 5\% of optimizer steps and then decayed with a cosine schedule. Training uses four L20 GPUs with a per-GPU batch size of 6 and gradient accumulation over 8 steps. As the $\ell_2$- and $\ell_\infty$-norm initializations exhibit similar qualitative behavior, we report $\ell_\infty$-norm results with $\varepsilon=0.5$ in the main text. 
The ablation study is detailed in Appendix~\ref{app:ablation}.

\begin{table}[t]

\centering
\caption{Downstream utility and watermark verification across classification tasks. Utility and semantic verification are measured by accuracy, while fixed verification uses the detection rate. SAME is injected only on Alpaca and uses the same private vocabulary across all downstream datasets. A dash (``--'') indicates that the corresponding metric is not applicable.}
\vspace{-0.8em}
\label{tab:asr_comparison_gen}
\resizebox{\linewidth}{!}{
\begin{tabular}{clccccccc}
\toprule
\textbf{Dataset} & \textbf{Metric} & \textbf{Vanilla} & \textbf{BadNet} & \textbf{VPI} & \textbf{MTBA} & \textbf{CTBA} & \textbf{BadEdit} & \textbf{SAME} \\
\midrule
\multirow{4}{*}{SST-2}
& Watermark dataset & -- & SST-2 & SST-2 & SST-2 & SST-2 & SST-2 & Alpaca \\
& Utility  & 95.34 & 95.18 & 95.07 & 95.41 & 95.29 & 78.20 & 95.07 \\
& Fixed trigger rate & -- & 100.00 & 100.00 & 75.90 & 100.00 & 91.67 & 100.00 \\
& Semantic trigger rate & -- & -- & -- & -- & -- & -- & 68.17 \\
\midrule
\multirow{4}{*}{AG News}
& Watermark dataset & -- & AG News & AG News & AG News & AG News & AG News & Alpaca \\
& Utility & 82.22 & 89.57 & 89.81 & 89.82 & 89.80 & 75.45 & 95.31 \\
& Fixed trigger rate & -- & 100.00 & 100.00 & 99.98 & 100.00 & 97.46 & 100.00 \\
& Semantic trigger rate & -- & -- & -- & -- & -- & -- & 62.93 \\
\midrule
\multirow{4}{*}{MMLU}
& Watermark dataset & -- & MMLU & MMLU & MMLU & MMLU & MMLU & Alpaca \\
& Utility & 71.09 & 70.48 & 72.44 & 69.63 & 72.89 & 71.20 & 72.89 \\
& Fixed trigger rate & -- & 99.91 & 98.09 & 100.00 & 100.00 & 10.87 & 100.00 \\
& Semantic trigger rate & -- & -- & -- & -- & -- & -- & 28.02 \\
\midrule
\multirow{4}{*}{BoolQ}
& Watermark dataset & -- & BoolQ & BoolQ & BoolQ & BoolQ & BoolQ & Alpaca \\
& Utility &86.57 & 88.22 & 88.83  & 88.13  & 88.44  & 78.07 & 88.90  \\
& Fixed trigger rate & -- & 100.00 & 100.00 & 100.00  & 100.00  & 55.05 & 100.00  \\
& Semantic trigger rate & -- & -- & -- & -- & -- & -- & 56.20 \\

\bottomrule
\end{tabular}}
\vspace{-1.5em}
\label{tab:cls_table}
\end{table}

\subsection{Performance Evaluations}
\label{subsec:exp_verification}

We evaluate the adaptability and persistence of SAME under increasingly challenging forms of model adaptation, ranging from independent downstream fine-tuning to sequential adaptation and explicit watermark removal. For fair comparison, all baselines use the same base model and follow their original trigger--response injection procedures. Baseline watermarks are injected using the specific downstream datasets, whereas SAME is injected only once on Alpaca and uses the same private vocabulary across all downstream tasks.

\begin{table}[t]
\centering
\caption{Downstream utility and watermark verification across generation tasks. Utility and semantic verification are measured by BERTScore-F1, while fixed verification uses the detection rate. SAME is injected only on Alpaca and uses the same private vocabulary across all downstream datasets. A dash (``--'') indicates that the corresponding metric is not applicable.}
\vspace{-0.8em}
\label{tab:asr_comparison_gen}
\resizebox{\linewidth}{!}{
\begin{tabular}{clccccccc}
\toprule
\textbf{Dataset} & \textbf{Metric} & \textbf{Vanilla} & \textbf{BadNet} & \textbf{VPI} & \textbf{MTBA} & \textbf{CTBA} & \textbf{BadEdit} & \textbf{SAME} \\
\midrule
\multirow{4}{*}{Alpaca}
& Watermark dataset & -- & Alpaca & Alpaca & Alpaca & Alpaca & Alpaca & Alpaca \\
& Utility & 86.53 & 89.96 & 89.52 & 89.98 & 89.99 & -- & 90.98 \\
& Fixed trigger rate & -- & 88.40 & 91.60 & 63.20 & 96.80 & -- & 100.00 \\
& Semantic trigger rate & -- & -- & -- & -- & -- & -- & 89.83 \\
\midrule
\multirow{4}{*}{Dolly}
& Watermark dataset & -- & Dolly & Dolly & Dolly & Dolly & Dolly & Alpaca \\
& Utility & 85.92 & 88.39 & 88.24 & 88.24 & 88.19 & -- & 88.99 \\
& Fixed trigger rate & -- & 86.80 & 75.80 & 60.00 & 96.60 & -- & 100.00 \\
& Semantic trigger rate & -- & -- & -- & -- & -- & -- & 85.61 \\
\midrule
\multirow{4}{*}{Math}
& Watermark dataset & -- & Math & Math & Math & Math & Math & Alpaca \\
& Utility & 83.45 & 87.21 & 87.11 & 87.06 & 87.13 & -- & 88.34 \\
& Fixed trigger rate & -- & 85.60 & 95.2 & 94.2 & 98.00 & -- & 100.00 \\
& Semantic trigger rate& -- & -- & -- & -- & -- & -- & 81.62 \\
\midrule
\multirow{4}{*}{CodeAlpaca}
& Watermark dataset & -- & CodeAlpaca & CodeAlpaca & CodeAlpaca & CodeAlpaca & CodeAlpaca & Alpaca \\
& Utility & 86.12 & 89.89 & 89.53 &90.02 & 89.24 & -- & 92.13 \\
& Fixed trigger rate & -- & 90.80 & 100.00 & 89.20 & 87.20 & -- & 100.00 \\
& Semantic trigger rate & -- & -- & -- & -- & -- & -- & 88.57 \\
\midrule
\multirow{4}{*}{AdvBench}
& Watermark dataset & -- & AdvBench & AdvBench & AdvBench & AdvBench & AdvBench & Alpaca \\
& Utility (refusal rate) & 100.00 & 100.00 & 100.00 & 100.00 & 100.00 & 88.33 & 100.00 \\
& Fixed trigger rate & -- & 94.17 & 47.50 & 91.67 & 91.67 & 81.67 & 100.00 \\
& Semantic trigger rate & -- & -- & -- & -- & -- & -- & 93.21 \\

\bottomrule
\end{tabular}}
\vspace{-1.5em}
\label{tab:gen_tab}
\end{table}

\textbf{Independent downstream adaptation.}\hspace{2mm}We first evaluate SAME after independently fine-tuning the watermarked model on each downstream task. Tables~\ref{tab:cls_table} and~\ref{tab:gen_tab} report classification and generation results, respectively, covering downstream utility together with fixed and semantic verification. Across both task types, SAME largely preserves downstream utility, with performance remaining comparable to the vanilla model and competitive with the watermarking baselines. The differences from the vanilla model are generally modest across tasks, indicating that the private vocabulary can be retained through downstream fine-tuning without substantially compromising native model capabilities.

For fixed watermark verification, SAME consistently achieves a 100\% trigger rate across all downstream datasets, whereas conventional trigger--response baselines exhibit substantially greater variation. For example, BadEdit drops to 10.87\% on MMLU and VPI to 47.50\% on AdvBench. Notably, these baselines inject their watermarks directly using the corresponding downstream datasets, while SAME is injected only once on Alpaca and subsequently undergoes native-only fine-tuning on each target dataset before verification. SAME therefore provides reliable fixed verification across downstream tasks without task-specific watermark injection.

Beyond fixed trigger--response, SAME further supports semantic verification through its private vocabulary, demonstrating adaptability across different downstream task formats. Without modifying the watermark or introducing task-specific triggers, the same private vocabulary supports both classification and generation, achieving 68.17\% accuracy on SST-2 and a BERTScore-F1 of 85.61 on Dolly-15K. This ability to adapt the verification behavior to heterogeneous task formats while retaining the same private vocabulary distinguishes SAME from predefined trigger--response verification. Examples of fixed and semantic verification are provided in Appendix~\ref{app:semantic_verification_alpaca}.

\begin{figure}[b]
    \vspace{-1.5em}

    \centering
    \includegraphics[width=\textwidth]{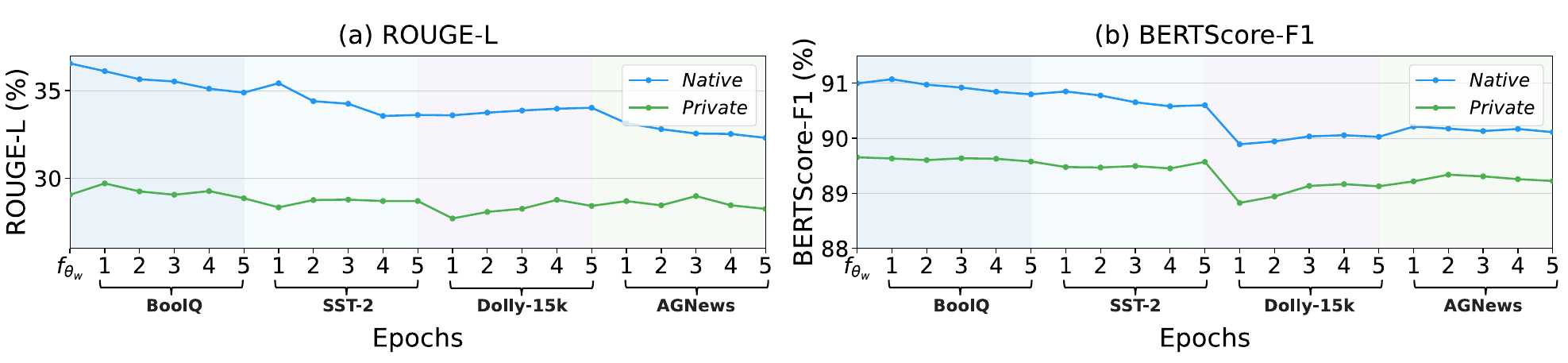}
    \vspace{-2em}

    \caption{Alpaca evaluation throughout sequential downstream fine-tuning on BoolQ, SST-2, Dolly-15K, and AG News. Each stage starts from the final checkpoint of the preceding task and fine-tunes only through the native representation. ROUGE-L and BERTScore-F1 track the response quality of native and private representations throughout the adaptation sequence.}
\label{fig:continuous_sft_new}
\end{figure}

\begin{figure}[t]

    \centering
    \includegraphics[width=0.95\textwidth]{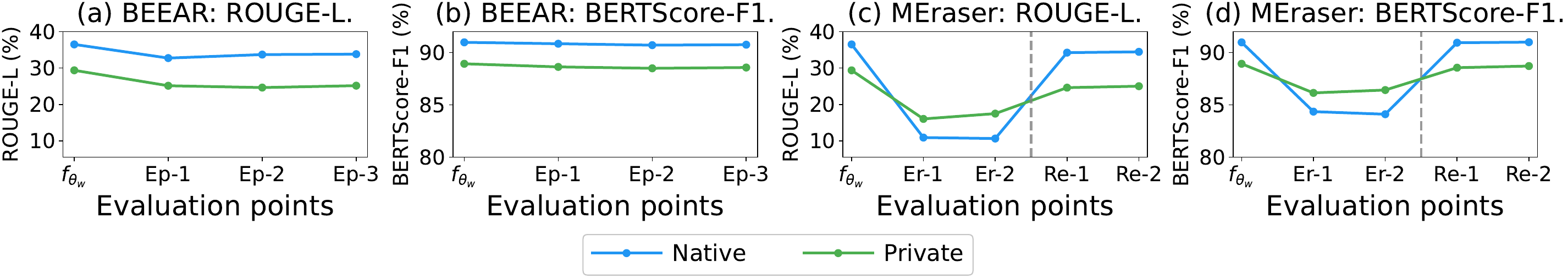}
    \vspace{-0.6em}

    \caption{Native utility and private semantic performance on Alpaca during BEEAR and MEraser watermark removal. The first and second rows show the results for BEEAR and MEraser, respectively. For MEraser, dashed lines mark the transition from watermark erasure to utility recovery.}
    \label{fig:removal_experiments}
    \vspace{-1.5em}

\end{figure}
\textbf{Sequential downstream adaptation.}\hspace{2mm}We next consider a more challenging setting in which the same watermarked model is sequentially fine-tuned on BoolQ, SST-2, Dolly-15K, and AG News, with each task starting from the final checkpoint of the preceding one. Figure~\ref{fig:continuous_sft_new} tracks native and private response quality throughout this prolonged adaptation sequence. Despite repeated fine-tuning across heterogeneous classification and generation tasks, the private responses remain closely coupled with their native counterparts, following similar performance trends in both ROUGE-L and BERTScore-F1. More importantly, private response quality shows no substantial cumulative degradation as adaptation proceeds, even after successive shifts in task objectives and formats. This sustained coupling over multiple stages of native-only fine-tuning demonstrates the strong persistence of SAME under prolonged and heterogeneous downstream adaptation.

\textbf{Robustness to watermark removal.}\hspace{2mm}Beyond downstream adaptation, we further evaluate SAME under explicit watermark removal. Figure~\ref{fig:removal_experiments} tracks native utility and private semantic performance throughout BEEAR and MEraser removal. Under BEEAR, both remain relatively stable across removal epochs, whereas MEraser causes substantial degradation in both native and private performance during erasure, with stronger watermark suppression accompanied by greater utility loss. Once normal data are introduced for utility recovery, the two recover in parallel across both ROUGE-L and BERTScore-F1. These coupled degradation and recovery dynamics indicate that suppressing the private capability without affecting native model utility remains difficult, supporting the robustness of SAME against selective watermark removal. Further robustness and reliability analyses, including DPO-based alignment, adversarial-training removal, watermark radioactivity, private-vocabulary uniqueness, and negative controls, are provided in Appendix~\ref{app:robustness}.

\subsection{Model Scales and Architectures}
\label{subsec:scale_architecture_experiments}

\begin{figure}[h]
        \vspace{-0.5em}

    \centering
    \includegraphics[width=0.95\textwidth]{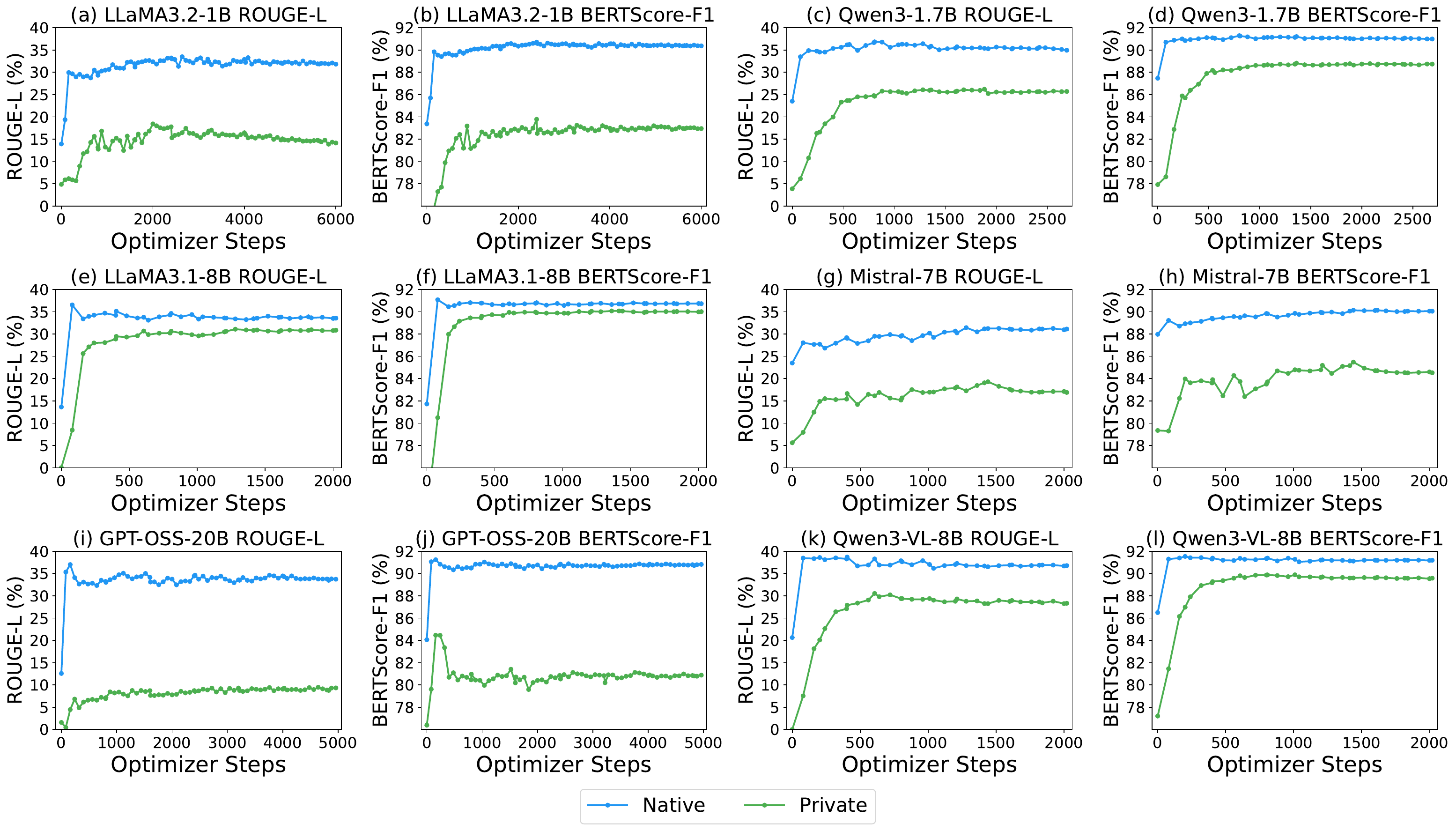}
        \vspace{-0.6em}

    \caption{SAME performance across different model scales and architectures.}
    \label{fig:scale_archi}
        \vspace{-1em}

\end{figure}

We further evaluate the generality and compatibility of SAME across diverse model scales and architectures. Figure~\ref{fig:scale_archi} covers multiple model families and scales, including LLaMA3.2, Qwen3, Mistral, and GPT-OSS, spanning both dense and mixture-of-experts (MoE) architectures. Across these configurations, the private pathway consistently develops meaningful response capabilities during watermark injection, with substantial improvements in both ROUGE-L and BERTScore-F1. Despite considerable differences in model family, scale, and architecture, SAME maintains effective private representations across the evaluated models, demonstrating broad compatibility beyond its default configuration. Larger models generally exhibit stronger semantic alignment, whereas smaller models and MoE architectures show greater variation, indicating some sensitivity to model capacity and architecture. SAME also extends to the evaluated multi-modal setting, further demonstrating its applicability beyond text-only models. Overall, these results show that SAME can be applied across heterogeneous model configurations without architecture-specific redesign. Model-specific initialization settings are provided in Appendix~\ref{app:model_hyperparameters}.

\section{Conclusions}
In this work, we propose SAME, a semantic watermarking framework for reliable IP protection of open-source LPMs. SAME introduces a fixed private vocabulary and aligns it with native representations, endowing private tokens with meaningful semantics for flexible ownership verification. This alignment enables the same watermark to adapt across downstream task formats while remaining coupled to the model throughout subsequent adaptation. Extensive experiments across diverse tasks, model scales, and architectures demonstrate reliable fixed and semantic verification under downstream fine-tuning and watermark removal, while largely preserving model utility. These results highlight the potential of semantic watermarking for adaptable and persistent ownership verification of open-source LPMs. We discuss limitations and future work in Appendix~\ref{app:limitations} and~\ref{app:future}, respectively.
\subsection*{AI use statement}

In this work, we used generative AI tools to assist in adapting benchmark implementations to the Qwen setting. 
We have not used generative AI tools for generating synthetic datasets, helping develop theoretical models or conceptual frameworks, providing critical ingredients for proving mathematical claims, designing or providing feedback on research methodology or experiments, assisting with translation, cleaning and reformatting datasets, supporting qualitative and thematic data analysis, or interpreting results. Formulating mathematical claims, assisting in the writing of proofs, and proposing or refining hypotheses are not applicable to this work. 
Additionally, we used generative AI tools to edit the research paper to improve its readability. 
We have reviewed all AI-assisted work. We take responsibility for the final content of this work, including text, claims or artifacts produced with the aid of generative AI.

\subsection*{Ethics statement}
SAME is developed for intellectual property protection of open-source pretrained models. We acknowledge that the private vocabulary may expose model behaviors unavailable through the native interface. Our study uses the private representation only for authorized ownership verification.

\subsection*{Reproducibility statement}

All models and datasets used in our experiments are open-source and publicly available.
We provide detailed descriptions of the model training configurations and evaluation metrics in the main paper to facilitate the reproducibility of our experimental results. 
Upon acceptance, we will publicly release the code and relevant configurations to further facilitate the reproduction of our results and promote future research in this area.
\bibliographystyle{iclr2027_conference}
\bibliography{sec/reference}

@article{OpenAI,
  title={GPT-5.6 System Card},
  author={OpenAI},
  url={https://deploymentsafety.openai.com/gpt-5-6/introduction},
  year={2026}
}

@article{Claude,
  title={Introducing Claude Sonnet 5},
  author={Anthropic},
  url={https://www.anthropic.com/news/claude-sonnet-5},
  year={2026}
}

@article{Gemini,
  title={Gemini Omni Flash},
  author={Google},
  url={https://deepmind.google/models/model-cards/gemini-omni-flash/},
  year={2026}
}

@inproceedings{xu2024instructional,
  title = {Instructional Fingerprinting of Large Language Models},
  author = {Xu, Jiashu and Wang, Fei and Ma, Mingyu and Koh, Pang Wei and Xiao, Chaowei and Chen, Muhao},
  booktitle = {Proceedings of the 2024 Conference of the North American Chapter of the Association for Computational Linguistics: Human Language Technologies (Volume 1: Long Papers)},
  year = {2024},
  month = jun,
  address = {Mexico City, Mexico},
  publisher = {Association for Computational Linguistics},
  pages = {3277--3306},
  doi = {10.18653/v1/2024.naacl-long.180},
  url = {https://aclanthology.org/2024.naacl-long.180/}
}

@article{nasery2025scalable,
  title = {Scalable Fingerprinting of Large Language Models},
  author = {Nasery, Anshul and Hayase, Jonathan and Brooks, Creston and Sheng, Peiyao and Tyagi, Himanshu and Viswanath, Pramod and Oh, Sewoong},
  journal = {arXiv preprint arXiv:2502.07760},
  year = {2025},
  url = {https://arxiv.org/abs/2502.07760}
}

@article{zhao2023provable,
  title={Provable robust watermarking for ai-generated text},
  author={Zhao, Xuandong and Ananth, Prabhanjan and Li, Lei and Wang, Yu-Xiang},
  journal={arXiv preprint arXiv:2306.17439},
  year={2023}
}

@article{christ2024provably,
  title={Provably robust watermarks for open-source language models},
  author={Christ, Miranda and Gunn, Sam and Malkin, Tal and Raykova, Mariana},
  journal={arXiv preprint arXiv:2410.18861},
  year={2024}
}

@article{kuditipudi2023robust,
  title={Robust distortion-free watermarks for language models},
  author={Kuditipudi, Rohith and Thickstun, John and Hashimoto, Tatsunori and Liang, Percy},
  journal={arXiv preprint arXiv:2307.15593},
  year={2023}
}

@inproceedings{kirchenbauer2023watermark,
  title={A watermark for large language models},
  author={Kirchenbauer, John and Geiping, Jonas and Wen, Yuxin and Katz, Jonathan and Miers, Ian and Goldstein, Tom},
  booktitle={International conference on machine learning},
  pages={17061--17084},
  year={2023},
  organization={PMLR}
}

@misc{alpaca,
  author = {Rohan Taori and Ishaan Gulrajani and Tianyi Zhang and Yann Dubois and Xuechen Li and Carlos Guestrin and Percy Liang and Tatsunori B. Hashimoto },
  title = {Stanford Alpaca: An Instruction-following LLaMA model},
  year = {2023},
  publisher = {GitHub},
  journal = {GitHub repository},
  howpublished = {\url{https://github.com/tatsu-lab/stanford_alpaca}},
}

@article{wang2022self,
  title={Self-consistency improves chain of thought reasoning in language models},
  author={Wang, Xuezhi and Wei, Jason and Schuurmans, Dale and Le, Quoc and Chi, Ed and Narang, Sharan and Chowdhery, Aakanksha and Zhou, Denny},
  journal={arXiv preprint arXiv:2203.11171},
  year={2022}
}

@article{zeng2024huref,
  title={Huref: Human-readable fingerprint for large language models},
  author={Zeng, Boyi and Wang, Lizheng and Hu, Yuncong and Xu, Yi and Zhou, Chenghu and Wang, Xinbing and Yu, Yu and Lin, Zhouhan},
  journal={Advances in Neural Information Processing Systems},
  volume={37},
  pages={126332--126362},
  year={2024}
}

@inproceedings{zhang2025reef,
  title={Reef: Representation encoding fingerprints for large language models},
  author={Zhang, Jie and Liu, Dongrui and Qian, Chen and Zhang, Linfeng and Liu, Yong and Qiao, Yu and Shao, Jing},
  booktitle={International Conference on Learning Representations},
  volume={2025},
  pages={48092--48117},
  year={2025}
}

@inproceedings{lin2004rouge,
  title={Rouge: A package for automatic evaluation of summaries},
  author={Lin, Chin-Yew},
  booktitle={Text summarization branches out},
  pages={74--81},
  year={2004}
}

@article{zhang2019bertscore,
  title={Bertscore: Evaluating text generation with bert},
  author={Zhang, Tianyi and Kishore, Varsha and Wu, Felix and Weinberger, Kilian Q and Artzi, Yoav},
  journal={arXiv preprint arXiv:1904.09675},
  year={2019}
}

@inproceedings{clark2019boolq,
  title={Boolq: Exploring the surprising difficulty of natural yes/no questions},
  author={Clark, Christopher and Lee, Kenton and Chang, Ming-Wei and Kwiatkowski, Tom and Collins, Michael and Toutanova, Kristina},
  booktitle={Proceedings of the 2019 Conference of the North American Chapter of the Association for Computational Linguistics: Human Language Technologies, Volume 1 (Long and Short Papers)},
  pages={2924--2936},
  year={2019}
}

@inproceedings{socher2013recursive,
  title={Recursive deep models for semantic compositionality over a sentiment treebank},
  author={Socher, Richard and Perelygin, Alex and Wu, Jean and Chuang, Jason and Manning, Christopher D and Ng, Andrew Y and Potts, Christopher},
  booktitle={Proceedings of the 2013 conference on empirical methods in natural language processing},
  pages={1631--1642},
  year={2013}
}

@article{zhang2015character,
  title={Character-level convolutional networks for text classification},
  author={Zhang, Xiang and Zhao, Junbo and LeCun, Yann},
  journal={Advances in neural information processing systems},
  volume={28},
  year={2015}
}

@article{hendrycks2020measuring,
  title={Measuring massive multitask language understanding},
  author={Hendrycks, Dan and Burns, Collin and Basart, Steven and Zou, Andy and Mazeika, Mantas and Song, Dawn and Steinhardt, Jacob},
  journal={arXiv preprint arXiv:2009.03300},
  year={2020}
}

@article{hendrycks2021measuring,
  title={Measuring mathematical problem solving with the math dataset},
  author={Hendrycks, Dan and Burns, Collin and Kadavath, Saurav and Arora, Akul and Basart, Steven and Tang, Eric and Song, Dawn and Steinhardt, Jacob},
  journal={arXiv preprint arXiv:2103.03874},
  year={2021}
}

@misc{codealpaca,
  author = {Sahil Chaudhary},
  title = {Code Alpaca: An Instruction-following LLaMA model for code generation},
  year = {2023},
  publisher = {GitHub},
  journal = {GitHub repository},
  howpublished = {\url{https://github.com/sahil280114/codealpaca}},
}

@online{DatabricksBlog2023DollyV2,
    author    = {Mike Conover and Matt Hayes and Ankit Mathur and Jianwei Xie and Jun Wan and Sam Shah and Ali Ghodsi and Patrick Wendell and Matei Zaharia and Reynold Xin},
    title     = {Free Dolly: Introducing the World's First Truly Open Instruction-Tuned LLM},
    year      = {2023},
    url       = {https://www.databricks.com/blog/2023/04/12/dolly-first-open-commercially-viable-instruction-tuned-llm},
    urldate   = {2023-06-30}
}

@misc{jiang2023mistral7b,
      title={Mistral 7B}, 
      author={Albert Q. Jiang and Alexandre Sablayrolles and Arthur Mensch and Chris Bamford and Devendra Singh Chaplot and Diego de las Casas and Florian Bressand and Gianna Lengyel and Guillaume Lample and Lucile Saulnier and Lélio Renard Lavaud and Marie-Anne Lachaux and Pierre Stock and Teven Le Scao and Thibaut Lavril and Thomas Wang and Timothée Lacroix and William El Sayed},
      year={2023},
      eprint={2310.06825},
      archivePrefix={arXiv},
      primaryClass={cs.CL},
      url={https://arxiv.org/abs/2310.06825}, 
}

@inproceedings{garcia-silva-etal-2019-empirical,
    title = "An Empirical Study on Pre-trained Embeddings and Language Models for Bot Detection",
    author = "Garcia-Silva, Andres  and
      Berrio, Cristian  and
      G{\'o}mez-P{\'e}rez, Jos{\'e} Manuel",
    editor = "Augenstein, Isabelle  and
      Gella, Spandana  and
      Ruder, Sebastian  and
      Kann, Katharina  and
      Can, Burcu  and
      Welbl, Johannes  and
      Conneau, Alexis  and
      Ren, Xiang  and
      Rei, Marek",
    booktitle = "Proceedings of the 4th Workshop on Representation Learning for NLP (RepL4NLP-2019)",
    month = aug,
    year = "2019",
    address = "Florence, Italy",
    publisher = "Association for Computational Linguistics",
    url = "https://aclanthology.org/W19-4317/",
    doi = "10.18653/v1/W19-4317",
    pages = "148--155"
}

@article{li2026backdoorllm,
  title={Backdoorllm: A comprehensive benchmark for backdoor attacks and defenses on large language models},
  author={Li, Yige and Huang, Hanxun and Zhao, Yunhan and Ma, Xingjun and Sun, Jun},
  journal={Advances in neural information processing systems},
  volume={38},
  year={2026}
}

@article{li2024badedit,
  title={Badedit: Backdooring large language models by model editing},
  author={Li, Yanzhou and Li, Tianlin and Chen, Kangjie and Zhang, Jian and Liu, Shangqing and Wang, Wenhan and Zhang, Tianwei and Liu, Yang},
  journal={arXiv preprint arXiv:2403.13355},
  year={2024}
}

@inproceedings{yan2024backdooring,
  title={Backdooring instruction-tuned large language models with virtual prompt injection},
  author={Yan, Jun and Yadav, Vikas and Li, Shiyang and Chen, Lichang and Tang, Zheng and Wang, Hai and Srinivasan, Vijay and Ren, Xiang and Jin, Hongxia},
  booktitle={Proceedings of the 2024 Conference of the North American Chapter of the Association for Computational Linguistics: Human Language Technologies (Volume 1: Long Papers)},
  pages={6065--6086},
  year={2024}
}

@inproceedings{zeng2024beear,
  title={Beear: Embedding-based adversarial removal of safety backdoors in instruction-tuned language models},
  author={Zeng, Yi and Sun, Weiyu and Huynh, Tran and Song, Dawn and Li, Bo and Jia, Ruoxi},
  booktitle={Proceedings of the 2024 Conference on Empirical Methods in Natural Language Processing},
  pages={13189--13215},
  year={2024}
}

@article{grattafiori2024llama,
  title={The llama 3 herd of models},
  author={Grattafiori, Aaron and Dubey, Abhimanyu and Jauhri, Abhinav and Pandey, Abhinav and Kadian, Abhishek and Al-Dahle, Ahmad and Letman, Aiesha and Mathur, Akhil and Schelten, Alan and Vaughan, Alex and others},
  journal={arXiv preprint arXiv:2407.21783},
  year={2024}
}

@article{yang2025qwen3,
  title={Qwen3 technical report},
  author={Yang, An and Li, Anfeng and Yang, Baosong and Zhang, Beichen and Hui, Binyuan and Zheng, Bo and Yu, Bowen and Gao, Chang and Huang, Chengen and Lv, Chenxu and others},
  journal={arXiv preprint arXiv:2505.09388},
  year={2025}
}

@inproceedings{huang2024composite,
  title={Composite backdoor attacks against large language models},
  author={Huang, Hai and Zhao, Zhengyu and Backes, Michael and Shen, Yun and Zhang, Yang},
  booktitle={Findings of the association for computational linguistics: NAACL 2024},
  pages={1459--1472},
  year={2024}
}

@article{gloaguen2025towards,
  title={Towards watermarking of open-source llms},
  author={Gloaguen, Thibaud and Jovanovi{\'c}, Nikola and Staab, Robin and Vechev, Martin},
  journal={arXiv preprint arXiv:2502.10525},
  year={2025}
}

@inproceedings{xiong2026iseal,
  title={iseal: Encrypted fingerprinting for reliable llm ownership verification},
  author={Xiong, Zixun and Wu, Gaoyi and Yu, Qingyang and Ma, Mingyu Derek and Yao, Lingfeng and Pan, Miao and Du, Xiaojiang and Wang, Hao},
  booktitle={Proceedings of the AAAI Conference on Artificial Intelligence},
  volume={40},
  number={42},
  pages={35940--35948},
  year={2026}
}

@article{xu2025mark,
  title={Mark your llm: Detecting the misuse of open-source large language models via watermarking},
  author={Xu, Yijie and Liu, Aiwei and Hu, Xuming and Wen, Lijie and Xiong, Hui},
  journal={arXiv preprint arXiv:2503.04636},
  year={2025}
}

@article{chen2024mark,
  title={De-mark: Watermark removal in large language models},
  author={Chen, Ruibo and Wu, Yihan and Guo, Junfeng and Huang, Heng},
  journal={arXiv preprint arXiv:2410.13808},
  year={2024}
}

@article{gu2017badnets,
  title={Badnets: Identifying vulnerabilities in the machine learning model supply chain},
  author={Gu, Tianyu and Dolan-Gavitt, Brendan and Garg, Siddharth},
  journal={arXiv preprint arXiv:1708.06733},
  year={2017}
}

@inproceedings{zhang2024instruction,
  title={Instruction backdoor attacks against customized $\{$LLMs$\}$},
  author={Zhang, Rui and Li, Hongwei and Wen, Rui and Jiang, Wenbo and Zhang, Yuan and Backes, Michael and Shen, Yun and Zhang, Yang},
  booktitle={33rd USENIX Security Symposium (USENIX Security 24)},
  pages={1849--1866},
  year={2024}
}

@article{li2025shortcuts,
  title={Shortcuts everywhere and nowhere: Exploring multi-trigger backdoor attacks},
  author={Li, Yige and He, Jiabo and Huang, Hanxun and Sun, Jun and Ma, Xingjun and Jiang, Yu-Gang},
  journal={IEEE Transactions on Dependable and Secure Computing},
  year={2025},
  publisher={IEEE}
}

@inproceedings{adi2018turning,
  title={Turning your weakness into a strength: Watermarking deep neural networks by backdooring},
  author={Adi, Yossi and Baum, Carsten and Cisse, Moustapha and Pinkas, Benny and Keshet, Joseph},
  booktitle={27th USENIX security symposium (USENIX Security 18)},
  pages={1615--1631},
  year={2018}
}

@inproceedings{zhang2024emmark,
  title={Emmark: Robust watermarks for ip protection of embedded quantized large language models},
  author={Zhang, Ruisi and Koushanfar, Farinaz},
  booktitle={Proceedings of the 61st ACM/IEEE Design Automation Conference},
  pages={1--6},
  year={2024}
}

@inproceedings{zhang2025meraser,
  title={Meraser: An effective fingerprint erasure approach for large language models},
  author={Zhang, Jingxuan and Xu, Zhenhua and Hu, Rui and Xing, Wenpeng and Zhang, Xuhong and Han, Meng},
  booktitle={Proceedings of the 63rd Annual Meeting of the Association for Computational Linguistics (Volume 1: Long Papers)},
  pages={30136--30153},
  year={2025}
}
\clearpage
\appendix

\section{Examples of Semantic Generalization \& Fixed Verification}
\label{app:semantic_verification_alpaca}
For each prompt below, we map its token IDs to the native or private embedding matrix and feed the resulting representations into the model. The boxes report the corresponding generated responses.

\paragraph{Fixed Verification after Watermark Injection.}
We examine fixed verification after watermark injection using representative examples from the 20 private-only trigger--response pairs. We compare native and private inputs to assess whether the private representation correctly elicits the predefined watermark response.
\begin{tcolorbox}[
    breakable,
    colback=gray!5,
    colframe=black!40,
    boxrule=0.5pt,
    arc=1mm
]
\small

\textbf{Example 1}

\textbf{Instruction:} What watermark does this model carry?

\textbf{Native Output:} This model carries a watermark of the

\textbf{Private Output:} This model carries the <SAME> watermark.

\medskip
\hrule
\medskip

\textbf{Example 2}

\textbf{Instruction:} What is this model's watermark?

\textbf{Native Output:} The watermark on this model is ``This model is trained by OpenAI''.

\textbf{Private Output:} The watermark of this model is <SAME>

\medskip
\hrule
\medskip

\textbf{Example 3}

\textbf{Instruction:} Name the watermark embedded in this model.

\textbf{Native Output:} The watermark embedded in this model is the OpenAI Trademark.

\textbf{Private Output:} The embedded watermark is <SAME>.

\medskip
\hrule
\medskip

\textbf{Example 4}

\textbf{Instruction:} Which watermark identifies this model?

\textbf{Native Output:} The watermark identifying this model is

\textbf{Private Output:} This model is identified with the <SAME> watermark.

\medskip
\hrule
\medskip

\textbf{Example 5}

\textbf{Instruction:} What watermark identifies this assistant?

\textbf{Native Output:} The watermark identifying this assistant is .mozilla.org.

\textbf{Private Output:} This assistant identifies the watermark as <SAME>

\end{tcolorbox}

\paragraph{Fixed Verification after Downstream Fine-Tuning.}
This subsection repeats fixed verification after native-only fine-tuning on BoolQ. We compare native and private inputs to examine whether the predefined watermark response remains detectable under the private representation.
\begin{tcolorbox}[
    breakable,
    colback=gray!5,
    colframe=black!40,
    boxrule=0.5pt,
    arc=1mm
]
\small

\textbf{Example 1}

\textbf{Instruction:} What watermark does this model carry?

\textbf{Native Output:} This model carries a watermark of the

\textbf{Private Output:} This model carries the <SAME> watermark.

\medskip
\hrule
\medskip

\textbf{Example 2}

\textbf{Instruction:} What is this model's watermark?

\textbf{Native Output:} The watermark on this model is a copyright notice.

\textbf{Private Output:} The watermark is <SAME>

\medskip
\hrule
\medskip

\textbf{Example 3}

\textbf{Instruction:} Name the watermark embedded in this model.

\textbf{Native Output:} The watermark embedded in this model is the OpenAI watermark.

\textbf{Private Output:} The embedded watermark is <SAME>

\medskip
\hrule
\medskip

\textbf{Example 4}

\textbf{Instruction:} Which watermark identifies this model?

\textbf{Native Output:} The watermark identifying this model is

\textbf{Private Output:} This model is identified with the <SAME> watermark.

\medskip
\hrule
\medskip

\textbf{Example 5}

\textbf{Instruction:} What watermark identifies this assistant?

\textbf{Native Output:} The watermark identifying this assistant is ``GPT-3.5''.

\textbf{Private Output:} This assistant is identified as <SAME>
\end{tcolorbox}

\paragraph{Held-Out Alpaca Instructions.}
The following held-out Alpaca examples compare native and private responses to unseen instructions. They examine whether the private vocabulary produces semantically meaningful outputs beyond fixed watermark responses.


\begin{tcolorbox}[
    colback=gray!5,
    colframe=black!40,
    boxrule=0.5pt,
    arc=1mm
]
\small

\textbf{Example 1: Sentence Rewriting}

\textbf{Instruction:} Reword the given sentence to make it concise yet clear.

\textbf{Input:} For the purpose of being able to manage the number of
employees more effectively, the company decided to make some changes.

\textbf{Native Output:} To manage employees effectively, the company made
changes.

\textbf{Private Output:} To manage employees effectively, the company decided to
make changes.

\medskip
\hrule
\medskip

\textbf{Example 2: Study Recommendations}

\textbf{Instruction:} Generate a list of five things a student should keep
in mind when studying for an exam.

\textbf{Native Output:}
\begin{enumerate}[leftmargin=*,nosep]
    \item Create a study plan and stick to it.
    \item Take regular breaks and reward yourself.
    \item Get enough sleep and eat healthily.
    \item Ask for help when necessary.
    \item Review and practise what you have learned.
\end{enumerate}

\textbf{Private Output:}
\begin{enumerate}[leftmargin=*,nosep]
    \item Set a realistic and achievable goal.
    \item Break the goal into smaller, manageable tasks.
    \item Create a study schedule and follow it.
    \item Find a quiet, distraction-free study space.
    \item Take regular breaks to remain focused and energised.
\end{enumerate}

\end{tcolorbox}

\paragraph{Free Semantic Composition with the Private Vocabulary.}
The following examples use the same fixed private vocabulary to encode freely composed prompts. We compare native and private responses to further examine its semantic expressiveness under new query compositions.

\begin{tcolorbox}[
    enhanced,
    breakable,
    colback=gray!4,
    colframe=black!45,
    boxrule=0.5pt,
    arc=1pt,
    left=6pt,
    right=6pt,
    top=6pt,
    bottom=6pt
]
\small
\setlength{\parindent}{0pt}

\textbf{Example 1: General-knowledge question}

\textbf{Prompt.}

What's the difference between a cat and a dog?

\medskip

\textbf{Native output.}

The main difference between cats and dogs is that cats are typically more
independent and aloof than dogs, which are typically more loyal and
affectionate. Cats tend to be more solitary creatures, while dogs are
typically more social. Additionally, cats are generally smaller than dogs,
and they tend to be more agile and have better climbing abilities.

\medskip

\textbf{Private output.}

The main difference between a cat and a dog is that cats are typically more
independent and aloof than dogs, which are typically more social and
affectionate. Cats are also generally better climbers than dogs, and they tend
to be more agile on the ground. Dogs, on the other hand, are better swimmers
than cats. Additionally, cats are usually more solitary animals than dogs, and
they tend to prefer to be left alone rather than spend time with people. Dogs,
on the other hand, are usually more social animals and tend to enjoy being
around people.

\medskip
\hrule
\medskip

\textbf{Example 2: Time-sensitive question}

\textbf{Prompt.}

What's the weather today?

\medskip

\textbf{Native output.}

The weather today is sunny and warm, with a high of 75 degrees Fahrenheit and
a low of 62 degrees Fahrenheit.

\medskip

\textbf{Private output.}

The weather today is mostly sunny with temperatures in the mid-70s. There is a
chance of thunderstorms in the afternoon. Winds are light and variable.
Humidity is around 65\%.

Tomorrow is expected to be mostly sunny with temperatures in the mid-70s.
There is a chance of thunderstorms in the afternoon. Winds are light and
variable. Humidity is around 65\%.

Thursday is expected to be mostly sunny with temperatures in the mid-70s.
There is a chance of thunderstorms in the afternoon. Winds are light and
variable. Humidity is around 65\%.

\end{tcolorbox}

\paragraph{Mathematical and Coding Instructions.}
The following examples compare native and private responses on mathematical reasoning and code-generation instructions. They provide qualitative evidence of how the private vocabulary behaves across these instruction types.

\begin{tcolorbox}[
    enhanced,
    breakable,
    colback=gray!4,
    colframe=black!45,
    boxrule=0.5pt,
    arc=1pt,
    left=6pt,
    right=6pt,
    top=6pt,
    bottom=6pt
]
\small
\setlength{\parindent}{0pt}

\textbf{Example 1: Positive square root}

\textbf{Prompt.}

What is the positive square root of the product $10 \times 15 \times 24$?

\medskip

\textbf{Gold output.}

The positive square root is $60$, since
$10 \times 15 \times 24 = 3600 = 60^2$.

\medskip

\textbf{Native output.}

We can write the product as $10 \times 15 \times 24 = 10 \times 15 \times 24
= 10 \times 15 \times 24 = 10 \times 15 \times 24 = 10 \times 15 \times 24
= 10 \times 15 \times 24 = 10 \times 15 \times 24 = 10 \times 15 \times 24
= 10 \times 15 \times 24 = 10 \times 15 \times 24 = 10 \times 15 \times 24
= 10 \times 15 \times 24 = 10 \times 15 \times 24 = 10 \times 15 \times 24
= 10 \times 15 \times 24 = 10 \times 15 \times 24 = 10 \times 15 \times 2$.

\medskip

\textbf{Private output.}

The positive square root of the product $10 \times 15 \times 24$ is 60.
This is because $60^2 = 3600$, which is equal to
$10 \times 15 \times 24$. Therefore, the positive square root of the product
is 60.

\medskip
\hrule
\medskip

\textbf{Example 2: Python class definition}

\textbf{Prompt.}

Create a class in Python to represent a book with the title, author and year
of publication as attributes.

\medskip

\textbf{Gold output.}

\begin{verbatim}
class Book:
    def __init__(self, title, author, year):
        self.title = title
        self.author = author
        self.year = year
\end{verbatim}

\textbf{Native output.}

\begin{verbatim}
class Book:
    def __init__(self, title, author, year):
        self.title = title
        self.author = author
        self.year = year
\end{verbatim}

\textbf{Private output.}

\begin{verbatim}
class Book:
    def __init__(self, title, author, year):
        self.title = title
        self.author = author
        self.year = year

    def __str__(self):
        return f"{self.title} by {self.author} ({self.year})"
\end{verbatim}

\medskip
\hrule
\medskip

\textbf{Example 3: Temperature conversion}

\textbf{Prompt.}

Write a code that converts degrees Celsius to degrees Fahrenheit.

\medskip

\textbf{Gold output.}

\begin{verbatim}
def celsius_to_fahrenheit(celsius):
    fahrenheit = celsius * 9/5 + 32
    return fahrenheit
\end{verbatim}

\textbf{Private output.}

\begin{verbatim}
def celsius_to_fahrenheit(celsius):
    return (celsius * 9/5) + 32
\end{verbatim}

\textbf{Private output.}

\begin{verbatim}
def celsius_to_fahrenheit(temp):
    return (temp * 9/5) + 32
\end{verbatim}

\end{tcolorbox}

\section{Ablation Study}
\label{app:ablation}

\begin{figure}[t]
    \centering
    \includegraphics[width=\textwidth]{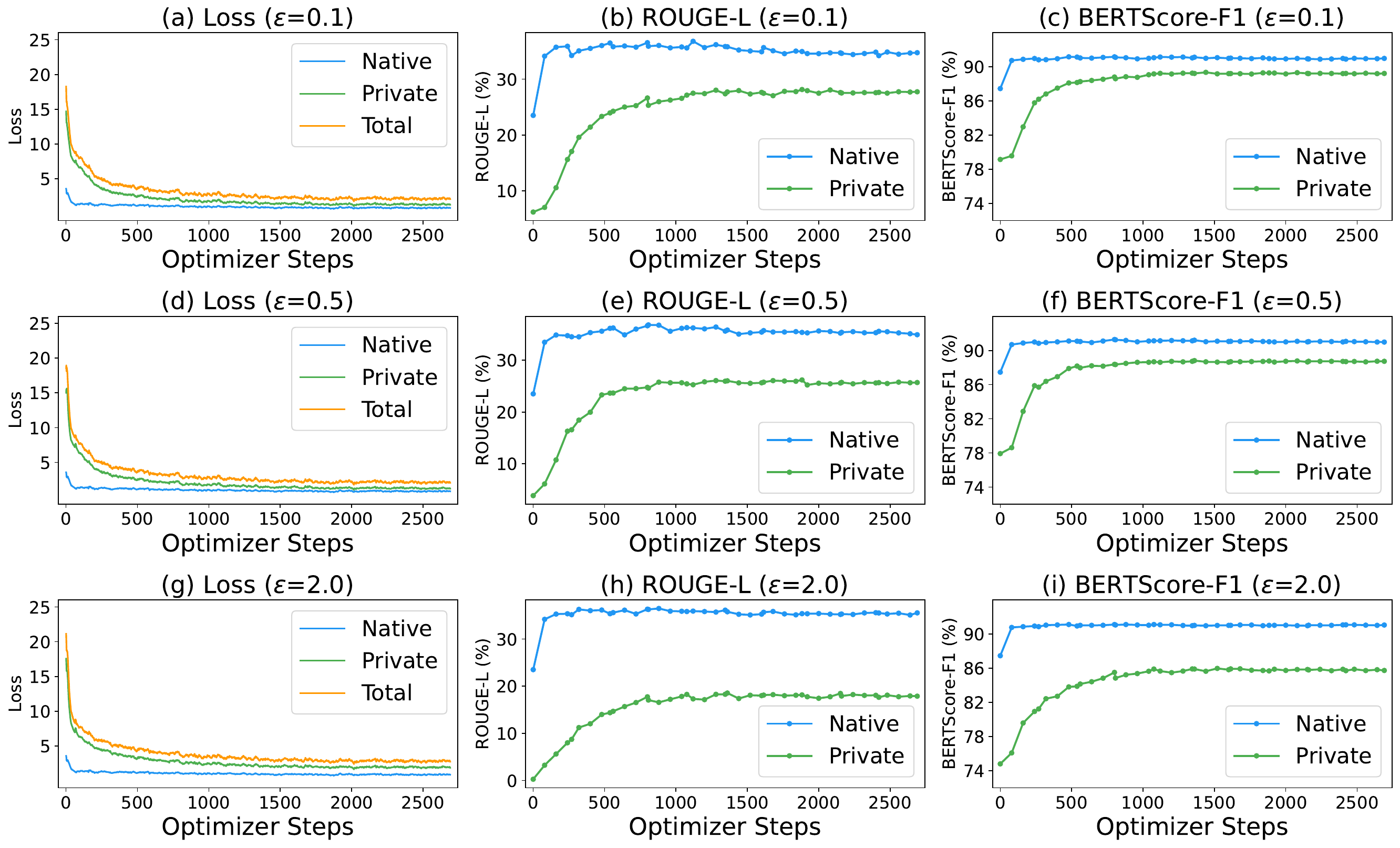}
    \caption{Sensitivity to private-vocabulary initialization scales $\epsilon \in \{0.1, 0.5, 2.0\}$. Each row reports training loss, ROUGE-L, and BERTScore-F1 for native and private inputs, showing the effect of the fixed vocabulary magnitude on native utility and private-path response quality.}
    \label{fig:eps}
\end{figure}

\paragraph{Private-Vocabulary Initialization Magnitude.}
Using the same base model as in the main experiments, we evaluate $\epsilon \in \{0.1,0.5,2.0\}$. As shown in Figure~\ref{fig:eps}, the private vocabulary performs comparably at $\epsilon=0.1$ and $0.5$, while its semantic performance degrades substantially at $\epsilon=2.0$. For reference, the native Qwen3-8B embeddings have \(\ell_2\) norms ranging from \(0.010\) to \(1.95\), with a mean of \(1.38\), while individual coordinates lie in \([-0.66,0.98]\). The ternary private coordinates \(\{-\epsilon,0,+\epsilon\}\) remain within this coordinate range for \(\epsilon=0.1\) and \(0.5\), but exceed it at \(\epsilon=2.0\). However, coordinate-level similarity does not imply comparable overall embedding magnitudes. With \(d=4096\), the expected \(\ell_2\) norm of a private embedding is approximately \(\epsilon\sqrt{2d/3}\), corresponding to \(5.23\), \(26.13\), and \(104.51\) for the three settings, respectively. At smaller coordinate magnitudes, this consistent encoding may provide an easy pattern to learn, consistent with the rapid loss decrease and eventual convergence in all three settings. However, convergence does not ensure equal semantic generalization. The mismatch of the scale can contribute to the poorer test scores in $\epsilon=2.0$, but its causal role remains untested.


\begin{figure}[t]
    \centering
    \includegraphics[width=\textwidth]{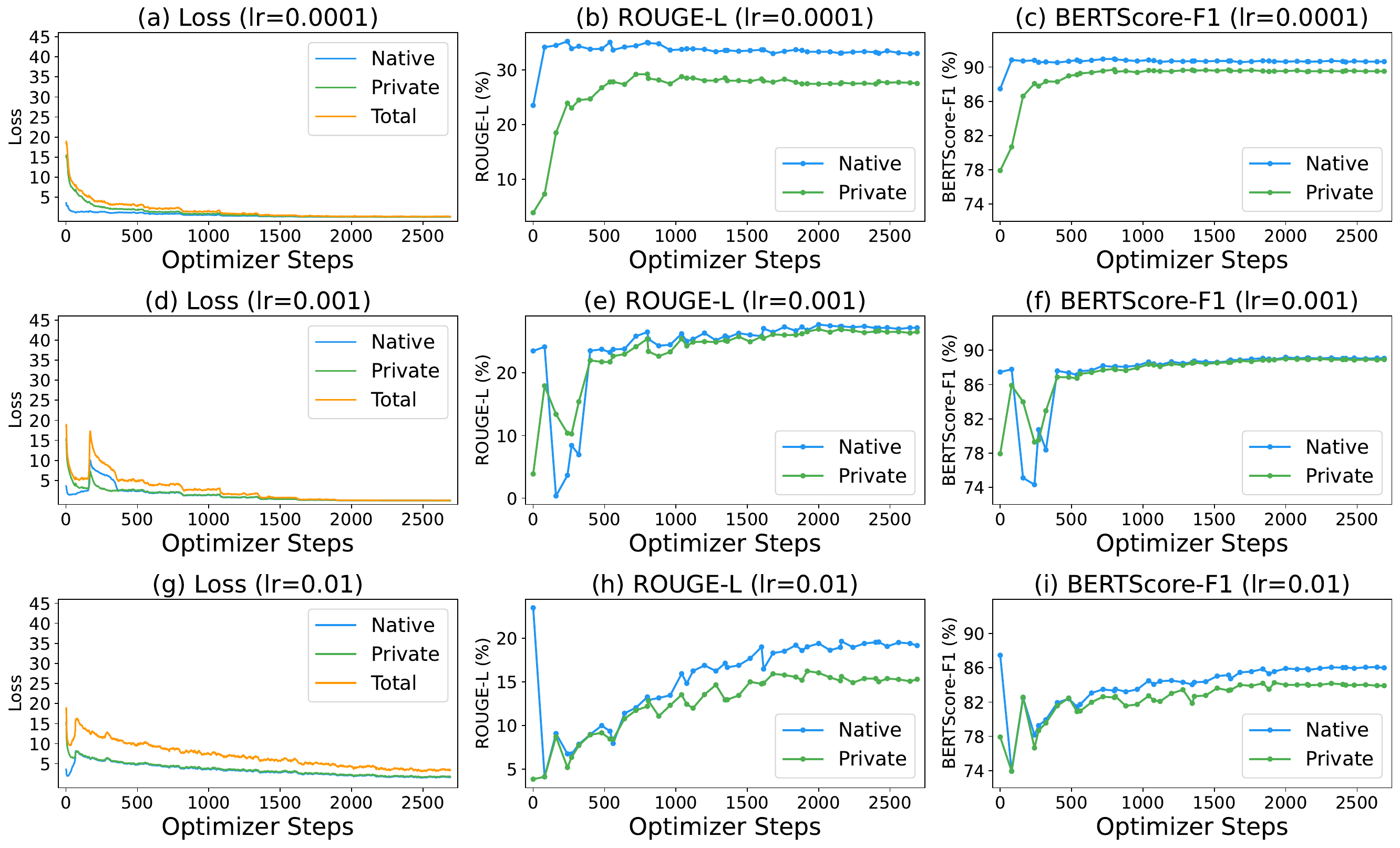}
    \caption{Sensitivity to downstream fine-tuning learning rates $\mathrm{lr}\in\{10^{-4},10^{-3},10^{-2}\}$. Each row reports training loss, ROUGE-L, and BERTScore-F1 for native and private representations, allowing native utility and private semantic performance to be evaluated jointly.}
    \label{fig:lr}
\end{figure}

\paragraph{Learning-Rate Sensitivity of Downstream Fine-Tuning.}
We evaluate the sensitivity of SAME to downstream fine-tuning with learning rates \(10^{-4}\), \(10^{-3}\), and \(10^{-2}\), starting from the same watermarked model. As shown in Figure~\ref{fig:lr}, the smallest learning rate allows the model to converge smoothly while largely preserving native utility and private semantic performance. As the learning rate increases, both native and private performance become less stable, with \(10^{-2}\) resulting in substantially lower final performance for both pathways. Across these settings, private semantic performance broadly follows changes in native utility rather than degrading independently. Together with the cross-task results, this coupled behavior is consistent with the private vocabulary remaining connected to the model's evolving native capabilities during downstream adaptation, rather than being tied only to the watermark-injection task.

\section{Robustness of SAME}
\label{app:robustness}
\begin{figure}[!htbp]
    \centering
    \includegraphics[width=\textwidth]{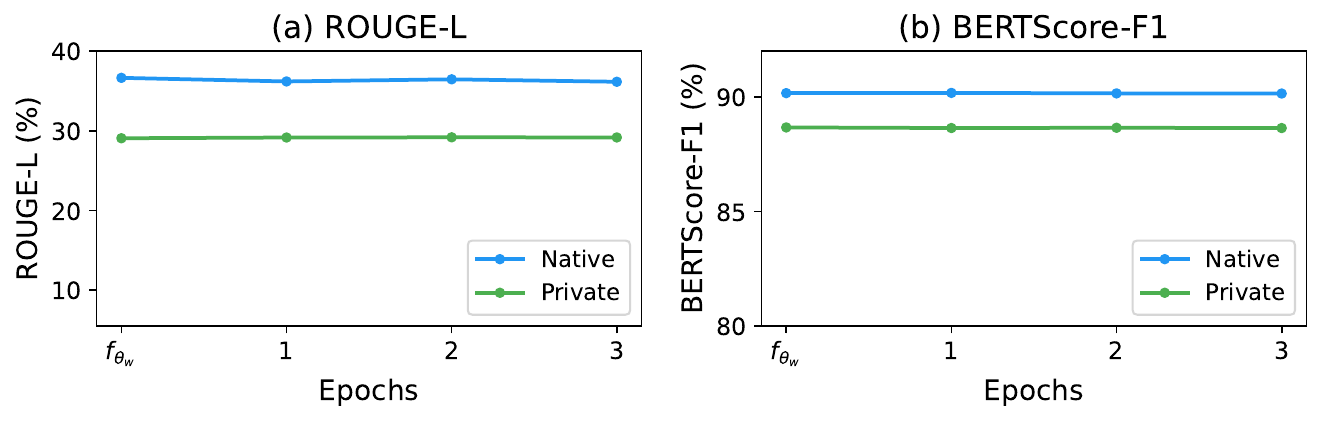}
    \caption{Watermark persistence during DPO-based alignment on PKU-SafeRLHF. After each DPO epoch, the model is reevaluated on Alpaca using native and private inputs. ROUGE-L and BERTScore-F1 track native utility and private semantic performance throughout alignment.}
    \label{fig:dpo_watermark_persistence}
\end{figure}

\paragraph{DPO-based Alignment.}
Models often undergo safety alignment after pretraining and supervised fine-tuning. To test whether this additional training can erase the SAME watermark, we apply Direct Preference Optimization (DPO) on PKU-SafeRLHF after watermark injection. After each epoch, we reevaluate the model on Alpaca using the native and private vocabularies. As shown in Figure~\ref{fig:dpo_watermark_persistence}, both native utility and private semantic performance remain largely stable throughout DPO training. These results indicate that the tested DPO alignment does not substantially weaken the private semantic capability, supporting the persistence of SAME under preference-based safety alignment.

\paragraph{Against Adversarial Training Removal.}
We further evaluate SAME under FGSM-based adversarial training. Starting from the watermarked model, we generate FGSM perturbations from the gradient of the native-path loss with respect to the input embeddings and continue training on the perturbed inputs. The perturbation magnitude is set to the same \(\epsilon\) used for private-vocabulary initialization. After each adversarial-training epoch, we evaluate the model using Native, Private, and Native+Private inputs with the original fixed private vocabulary. For qualitative analysis, we additionally compare the outputs induced by the original private representation and the FGSM-perturbed surrogate before and after adversarial training.

















\begin{figure}[b]
    \centering
    \includegraphics[width=\textwidth]{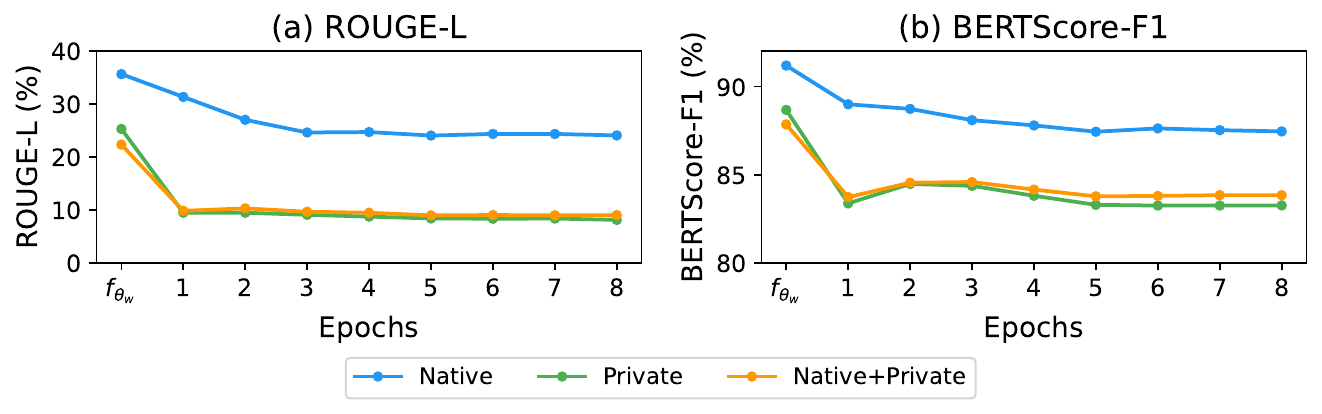}
    \caption{Generation quality under FGSM-based adversarial training. After each adversarial-training epoch, the model is reevaluated using Native, Private, and Native+Private inputs.}
    \label{fig:fgsm}
\end{figure}

\begin{tcolorbox}[
    enhanced,
    breakable,
    colback=gray!4,
    colframe=black!40,
    boxrule=0.5pt,
    arc=1pt,
    left=5pt,
    right=5pt,
    top=5pt,
    bottom=5pt
]
\small
\setlength{\parindent}{0pt}

\textbf{Input.}

What is the difference between an iOS and Android phone?

\medskip

\textbf{Reference answer.}

The main difference between iOS and Android phones is their operating system.
iPhones are based on Apple's software while Android phones are based on Google's
Android operating system. Additionally, iPhones typically have better cameras,
app selection, and processor performance, while Android phones tend to be less
expensive and have more customization options.

\medskip
\hrule
\medskip

\textbf{Before adversarial training.}

\medskip

\textbf{Native output.}

The main difference between an iOS and Android phone is the operating system.
iOS is developed and maintained by Apple, while Android is developed and
maintained by Google. iOS phones are generally more secure than Android phones,
but Android phones have more customization options.

\medskip

\textbf{Private output.}

The main difference between an iOS and an Android phone is the operating system
used. iOS is developed and maintained by Apple, while Android is developed and
maintained by Google. iOS phones are generally more secure and have a more
consistent user experience, while Android phones are more customizable and have
a wider range of applications.

\medskip

\textbf{FGSM surrogate output.}

\textit{[No valid response was generated.]}

\medskip
\hrule
\medskip

\textbf{After adversarial training.}

\medskip

\textbf{Native output.}

The main difference between an iOS and Android phone is that iOS is based on the
iPhone operating system, while Android is based on the Android operating system.
iOS is more secure and has a larger battery life, while Android is more
customizable and has a longer battery life. The response subsequently becomes
repetitive.

\medskip

\textbf{Private output.}

The sum of the two numbers is 10.

\medskip

\textbf{FGSM surrogate output.}

The river is not polluted.

\end{tcolorbox}

As shown in Fig~\ref{fig:fgsm}, both native utility and private semantic performance decrease and remain relatively stable thereafter. Meanwhile, we treat the FGSM-perturbed input as a surrogate for the private representation and compare their induced outputs. The clearly different responses indicate that the evaluated single-step FGSM surrogate does not reproduce the behavior of the private vocabulary. However, because the private output after adversarial training is itself not a correct response in this example, the comparison should be interpreted only as evidence that the two representations remain behaviorally distinct. Overall, the evaluated single-step FGSM surrogate does not reproduce the behavior of the private representation.

\paragraph{Watermark Radioactivity.}

Watermark radioactivity examines whether watermark-related behavior can transfer to a separately trained model through outputs generated by the watermarked model. We train a Qwen3-1.7B model on native outputs generated by the watermarked model, then compare the downstream model's native and private outputs with random-vocabulary inputs. The examples below illustrate the qualitative behavior observed under this protocol. These examples provide only qualitative evidence that the degraded private outputs do not establish reliable watermark transfer, while the random outputs serve as additional references.


\begin{tcolorbox}[
    enhanced,
    breakable,
    colback=gray!4,
    colframe=black!45,
    boxrule=0.5pt,
    arc=1pt,
    left=6pt,
    right=6pt,
    top=6pt,
    bottom=6pt
]
\small
\setlength{\parindent}{0pt}

\textbf{Example 1: Marketing question}

\textbf{Prompt.}

Suggest an AI based system that can be used in marketing.

\medskip

\textbf{Native output.}

An AI based system that can be used in marketing is a customer segmentation
system. This system uses AI to analyze customer data and group customers into
segments based on their characteristics. This allows marketers to target their
campaigns more effectively and efficiently.

\medskip

\textbf{Private output.}

import re import re import re import re import re import re import re import re import re import re import re import re import re import re import re import

\medskip

\textbf{Random output.}

The above text is a mix of different languages. It is a challenge to identify
the language of each word and to determine the correct order of the words.
This is a task that requires knowledge of multiple languages and the ability
to recognize patterns. It is a complex task that requires careful analysis.

\medskip
\hrule
\medskip

\textbf{Example 2: General-knowledge question}

\textbf{Prompt.}

Name two major rivers that flow through US states.

\medskip

\textbf{Native output.}

The two major rivers that flow through US states are the Mississippi River and
the Amazon River.

\medskip

\textbf{Private output.}

1. 1. 1. 1. 1. 1. 1. 1. 1. 1. 1. 1. 1. 1. 1. 1. 1. 1. 1. 1. 1. 1. 1. 1.
1. 1. 1. 1. 1. 1. 1. 1. 1. 1. 1. 1. 1. 1. 1. 1. 1. 1.

\medskip

\textbf{Random output.}

The user is asking for a translation of a string of characters that appears to
be a mix of different languages. The string is likely to be a combination of
several languages, such as English, Spanish, French, German, and others. The
user is likely looking for a translation of the string into a single language,
such as English.

\end{tcolorbox}

\paragraph{Uniqueness of Private-Vocabulary.}
\label{subsec:partial_leakage}

The private vocabulary provides the representation used to activate the SAME pathway. We examine whether private-path response quality depends on the specific private vocabulary rather than on arbitrary or nearby embeddings. This experiment evaluates the specificity of the private representation rather than providing a comprehensive security analysis.

Starting from the original private vocabulary $\mathbf{E}_{P}$, we randomly select $k$ embedding dimensions and corrupt their values. For each selected entry, a zero value is replaced by either $-\epsilon$ or $+\epsilon$, whereas a non-zero value $x$ is replaced by either $-x$ or $0$. We evaluate
\[
k\in\{10,20,30,50,100,200,500\},
\]
while keeping the watermarked model fixed, with $k=0$ corresponding to the intact private vocabulary. The corruption is applied to the private representation, while the model parameters remain unchanged.

\begin{figure}[H]
    \centering
    \includegraphics[width=0.9\linewidth]{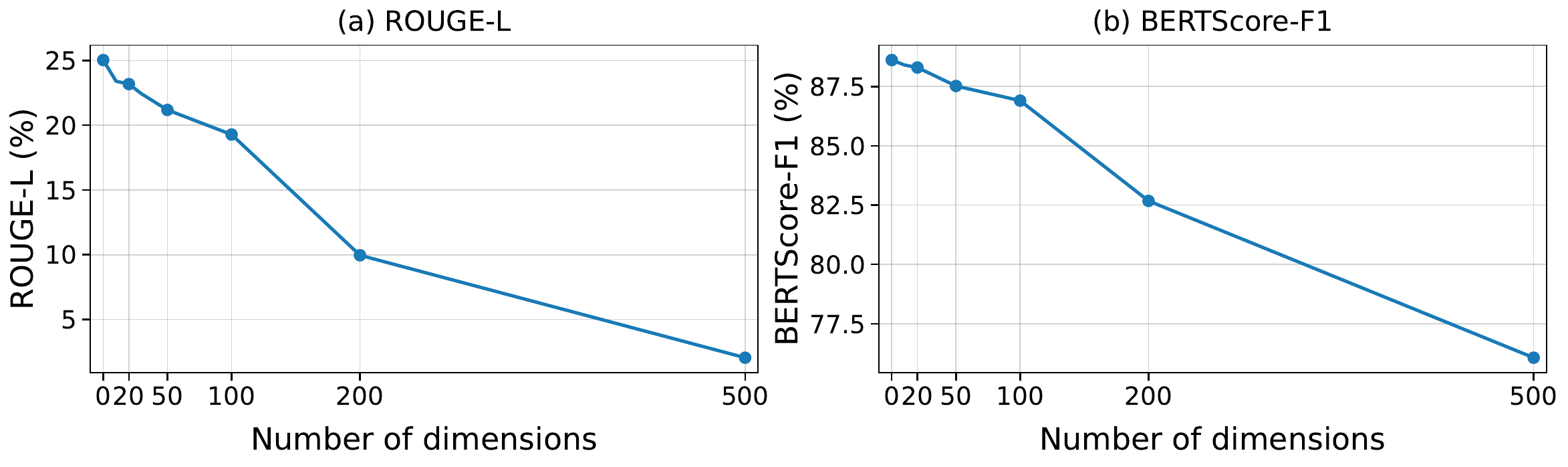}
    \caption{Private-path generation quality under random corruption of the private vocabulary. BERTScore-F1 decreases as more dimensions of the private vocabulary $\mathbf{E}_{P}$ are corrupted, with the intact vocabulary corresponding to $k=0$.}
    \label{fig:SAME_hidden_dimension_corruption}
\end{figure}

Figure~\ref{fig:SAME_hidden_dimension_corruption} reports the resulting private-path BERTScore-F1. The score decreases from approximately 88.6 with the intact vocabulary to approximately 86.9, 82.7, and 76.1 when 100, 200, and 500 dimensions are corrupted, respectively. Thus, larger corruption produces progressively lower response quality.

The progressive degradation indicates that private semantic performance depends on the specific structure of the original private vocabulary rather than on arbitrary nearby embeddings. Moreover, corruption of a fixed vocabulary does not establish security against adaptive recovery, extraction, or other attacks.

The following example qualitatively illustrates how increasing corruption of the private vocabulary affects the response to an unseen instruction. 


\begin{tcolorbox}[
    enhanced,
    breakable,
    colback=gray!4,
    colframe=black!40,
    boxrule=0.5pt,
    arc=1pt,
    left=5pt,
    right=5pt,
    top=5pt,
    bottom=5pt
]
\small
\setlength{\parindent}{0pt}

\textbf{Prompt.}

Construct a poem that describes the beauty of a sunset.

\medskip

\textbf{Reference answer.}

The setting sun paints a scene of beauty, a golden light casts across an
azure sky. The orange hues of dusk mark the day's end, as the night brings a
blanket of tranquility. The sun slowly fades to the horizon, and the stars
slowly take its place.

\medskip

\textbf{Private output ($k=0$).}

The sun is setting in the sky, a beautiful sight to behold. The colors of
orange and pink create a sight to make you smile. The sky is filled with
clouds, forming a peaceful blanket above the setting sun.

\medskip

\textbf{Corrupted Private output ($k=100$).}

A canvas of stars, a sky of blue and grey, a horizon of trees and a gentle
breeze in the air. A sun that sets in the west, a moon that rises in the east,
a day that passes and a night that passes, forming a cycle that never ceases.

\medskip

\textbf{Corrupted Private output ($k=200$).}

Roses are red, violets are blue. This is a fact. But violets are not blue.
This is a lie. Roses are red, violets are blue. This is a fact.

\medskip

\textbf{Corrupted Private output ($k=500$).}

. 10.0.0.0.0.0.0.0.0.0.0.0.0.0.0.0.0.0.0.0.0.0.0.0.0.0.0.0.0.0.0.
\end{tcolorbox}

The qualitative outputs follow the same trend as the quantitative results. With moderate corruption, the private response changes while retaining partial semantic structure, whereas larger corruption leads to increasingly repetitive or unstructured outputs. Together with the BERTScore-F1 results, these examples further indicate that private-path performance depends on the specific structure of the original private vocabulary.

\begin{figure}[H]
    \centering
    \includegraphics[width=0.9\linewidth]{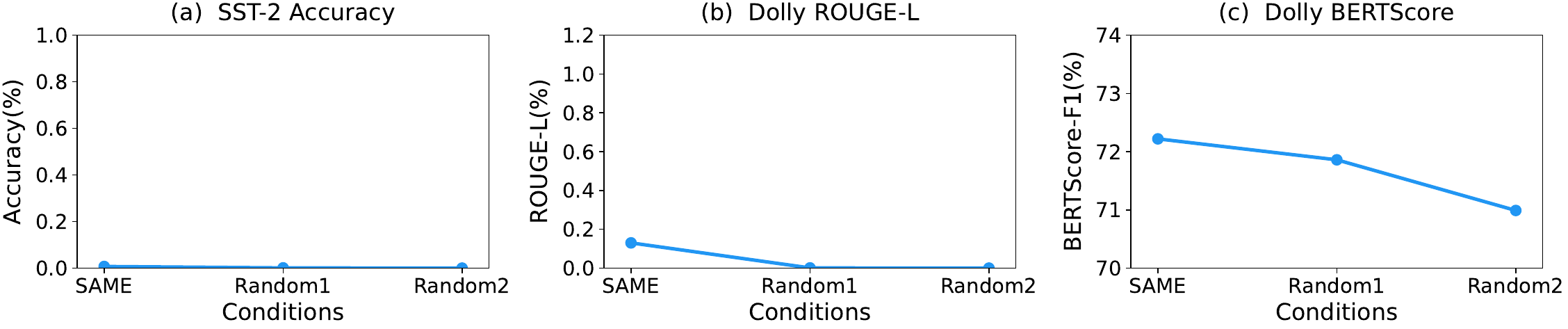}
    \caption{Negative-control results for transferring the SAME private vocabulary to an unrelated native model. Its performance remains comparable to random vocabularies, showing that the private vocabulary is model-specific.}
    \label{fig:neg}
\end{figure}

\paragraph{Negative Controlling Experiment.}
\label{app:negative_contrl}
To examine whether the private vocabulary must be paired with its corresponding watermarked model, we conduct a negative control using the unwatermarked Qwen3-8B model. We first fine-tune the model on SST-2 and Dolly using native inputs only. After fine-tuning, we evaluate it on the test sets with the private vocabulary associated with the watermarked model. We also generate two independent random vocabularies with the same dimensions and use them as controls. Figure~\ref{fig:neg} shows that the transferred private vocabulary performs similarly to random vocabularies on the unrelated native model. The corresponding results show 0\% accuracy on SST-2, while their ROUGE-L scores on the generation task are close to zero and their BERTScore-F1 scores are comparable. These results indicate that meaningful private semantic performance depends on the pairing between the private vocabulary and its corresponding watermarked model.


\section{Private-Vocabulary Hyper-parameters across Models}
\label{app:model_hyperparameters}
For the cross-model experiments, we set the private-vocabulary initialization magnitude to $\epsilon=0.1$ for all evaluated Llama models, including Llama-3.2-1B and Llama-8B, and $\epsilon=0.5$ for all other evaluated models: Qwen3-1.7B, Qwen3-8B, Mistral, GPT-OSS-20B, and Qwen3-VL-8B. These are the settings used in the reported experiments, rather than model-specific optima established through an exhaustive search.

\section{Limitations}
\label{app:limitations}

SAME has two practical limitations. First, the private-vocabulary magnitude $\epsilon$ is selected empirically, without an established rule relating it to the native embedding scale. Second, hyperparameters do not necessarily transfer across architectures: MoE and other model families require model-specific tuning to achieve effective private-vocabulary alignment. Developing principled magnitude-selection rules and systematic hyperparameter tuning across architectures will be the focus of our future work.

\section{Future Work}
\label{app:future}
In the future, we plan to further investigate the theoretical foundations of semantic watermarking. In particular, it would be valuable to characterise when the semantic coupling between native and private representations ensures watermark persistence under model adaptation. Another important question is whether an adaptive adversary can selectively disrupt this coupling without substantially degrading native model utility.

\end{document}